\pdfoutput=1

\documentclass[11pt]{article}
\usepackage{authblk}
\usepackage[]{acl}

\usepackage{times}
\usepackage{latexsym}

\definecolor{Latn_color}{HTML}{0000ff}
\definecolor{Cyrl_color}{HTML}{ff0000}
\definecolor{Hani_color}{HTML}{00A300}
\definecolor{Arab_color}{HTML}{ff8000}
\definecolor{Deva_color}{HTML}{ffcccc}
\definecolor{Grek_color}{HTML}{666633}
\definecolor{Hang_color}{HTML}{00ffff}
\definecolor{Thai_color}{HTML}{A30021}
\definecolor{Mlym_color}{HTML}{999966}
\definecolor{Hebrew_color}{HTML}{8000ff}

\usepackage{adjustbox}
\usepackage[T1]{fontenc}
\usepackage[T5]{fontenc}
\usepackage[utf8]{inputenc}
\usepackage{booktabs}
\usepackage[T2A,T1]{fontenc}
\usepackage[ukrainian,english]{babel}

\usepackage{microtype}
\usepackage{amsmath}
\usepackage{graphicx}
\usepackage{subfigure}
\usepackage{amsfonts}
\usepackage{multirow}
\usepackage{listings}
\usepackage{url}
\usepackage[ruled,linesnumbered]{algorithm2e}  
\usepackage{CJKutf8}
\usepackage{graphicx}
\usepackage{subfigure}
\usepackage{supertabular}
\usepackage{subfigure}
\usepackage{dcolumn,booktabs}
\usepackage{tikz}
\usepackage[right]{rotating}

\usepackage{inconsolata}

\title{\textsc{TransliCo}: A Contrastive Learning Framework to Address the Script Barrier in Multilingual Pretrained Language Models}

\author[]{\bf{Yihong Liu}$^{\text *}$}
\author[]{\bf{Chunlan Ma}$^{\text *}$}
\author[]{\bf{Haotian Ye}$^{\text *}$}
\author[]{\bf Hinrich Sch\"utze}

\affil{Center for Information and Language Processing, LMU Munich \\ Munich Center for Machine Learning (MCML)
 \protect\\ \texttt{\{yihong, chunlan, yehao\}@cis.lmu.de}}

\def\secref#1{\S\ref{sec:#1}}
\def\seclabel#1{\label{sec:#1}}

\newcounter{notecounter}
\newcommand{\enotesoff}{\long\gdef\enote##1##2{}}
\newcommand{\enoteson}{\long\gdef\enote##1##2{{
\stepcounter{notecounter}
{\large\bf
\hspace{1cm}\arabic{notecounter} $<<<$ ##1: ##2
$>>>$\hspace{1cm}}}}}
\enoteson
\enotesoff

\begin{document}
\maketitle

\def\thefootnote{*}\footnotetext{Equal contribution.}\def\thefootnote{\arabic{footnote}}
\def\frameworkname{\textsc{TransliCo}\xspace}
\def\modelname{\textsc{Furina}\xspace}
\def\modelnameIndi{\textsc{Furina}$_{\text{Indic}}$\xspace}

\begin{abstract}

The world's more than 7000 languages are written in
at least 293 scripts.\footnote{\url{https://worldswritingsystems.org/}} Due to various reasons, many closely related languages use different scripts, which poses a difficulty for multilingual pretrained language models (mPLMs) in learning crosslingual knowledge through lexical overlap.
As a consequence, mPLMs are faced with a script barrier:
representations from different scripts are located in
different subspaces, which can result in crosslingual
transfer involving languages of different scripts
performing suboptimally.
To address this problem, we propose \textbf{\frameworkname},
a framework that optimizes the Transliteration Contrastive
Modeling (TCM) objective to fine-tune an mPLM by contrasting sentences
in its training data and their transliterations in a unified
script (in our case Latin\footnote{Throughout this paper we use Latin to refer to the Latin script,
not the Latin language.}),
which enhances uniformity in the representation space for different scripts. 
Using Glot500-m \citep{imanigooghari-etal-2023-glot500}, an mPLM pretrained on over 500 languages, as our source model, we fine-tune it on a small portion (5\%) of its training data, and refer to the resulting model as \textbf{\textsc{Furina}}. We show that \textsc{Furina} not only better aligns representations from distinct scripts but also outperforms the original Glot500-m on various zero-shot crosslingual transfer tasks. Additionally, we achieve consistent improvement in a case study on the Indic group where the languages exhibit areal features but use different scripts. We make our code and models publicly available.\footnote{\url{https://github.com/cisnlp/TransliCo}}
\end{abstract}

\section{Introduction}

\begin{figure}
    \centering
    \includegraphics[width=0.5\textwidth]{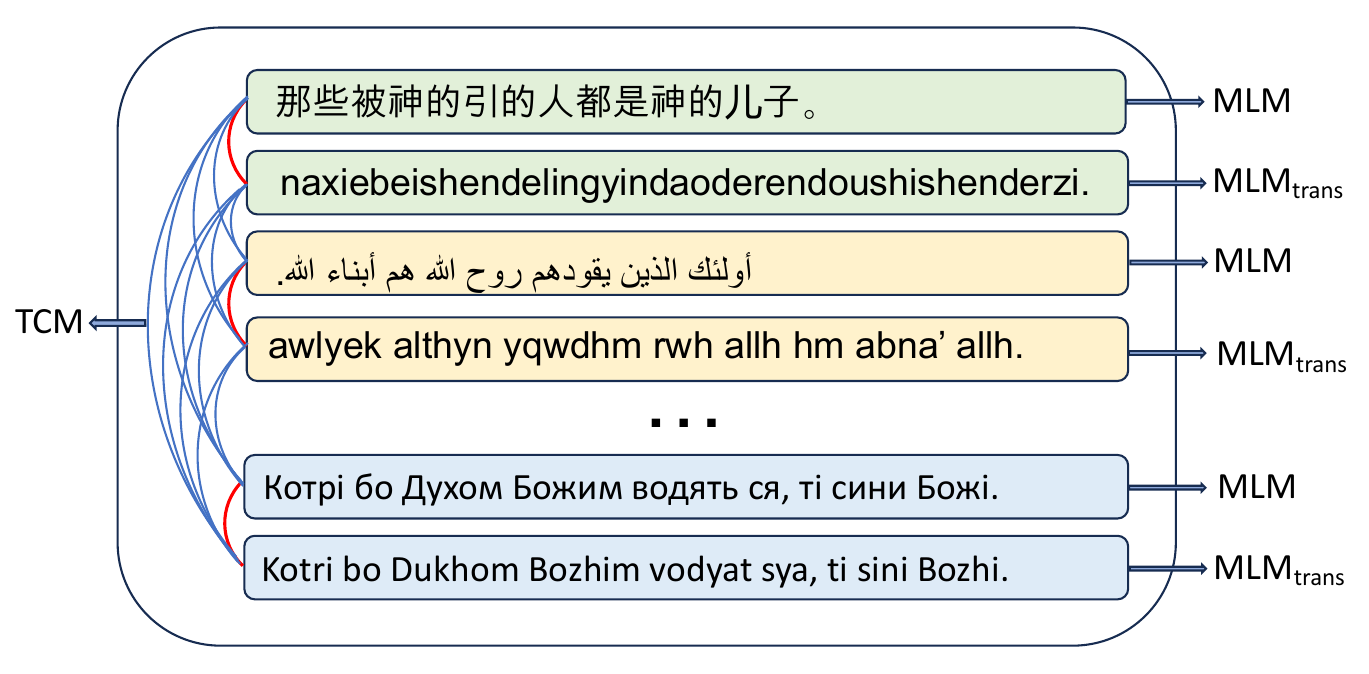}
    \caption{
    An illustration of applying \frameworkname to a single batch of data during fine-tuning. The training data is used by the two training objectives in \frameworkname: Masked Language Modeling (MLM) and Transliteration Contrastive Modeling (TCM). MLM is applied to both the original sentences and their Latin transliterations. TCM is used to learn better-aligned cross-script representations by contrasting the \underline{\textcolor{red}{positive pairs}} (paired data connected with red lines) against the \underline{\textcolor{blue}{negative pairs}} (the remaining samples connected with blue lines).}
    \label{fig:model_illustration}
\end{figure}

In recent years, mPLMs have made impressive progress in various crosslingual transfer tasks \citep{conneau-etal-2018-xnli,hu2020xtreme,liang-etal-2020-xglue}. Such achievement is mainly due to the availability of monolingual corpora of many languages \citep{costa2022no,adebara-etal-2023-serengeti,imanigooghari-etal-2023-glot500}, the amelioration of model architectures suitable for scaling up \citep{vaswani2017transformer,peng-etal-2023-rwkv}, as well as the advancement of self-supervised learning objectives \citep{devlin-etal-2019-bert,raffel2020t5,brown2020gpt}. Despite the fact that mPLMs present attractive performance in high-resource languages, those models often gain unsatisfactory results for low-resource languages, especially when the writing systems or \emph{scripts} are different from the transfer source languages \citep{muller-etal-2021-unseen}.

This undesired behavior is related to the script barrier
in the representation space, where different scripts
are located in different
subspaces \citep{wen-yi-mimno-2023-hyperpolyglot}. To tackle
this problem, transliteration or
romanization\footnote{Romanization is a specific type of
transliteration that involves converting non-Latin scripts
into the Latin script.} is leveraged in some recent
work \citep{dhamecha-etal-2021-role,muller-etal-2021-unseen,moosa-etal-2023-transliteration,
purkayastha-etal-2023-romanization}:
all languages from different scripts are converted into one common script
and the language model is pretrained or adapted with transliterated
data. For testing and inference, the queries also need to be
transliterated, as the model only supports one script, the
pretraining or adaptation script of the model.

However, this line of approaches presents two limitations. First, it does not break the script barrier,
rather, it circumvents it. The representations from different scripts are still not aligned. 
Second, for some tasks, e.g., question answering, it is
necessary to transliterate the response back to the original
script because we cannot assume that end users know the common script. Unfortunately, transliteration and transliterating back to the original script is not immune to information loss \citep{amrhein-sennrich-2020-romanization}. The romanized words in many languages, e.g., Chinese, Japanese, and Korean, can be converted to different words in their original scripts, which unfortunately leads to ambiguity. 

In this paper, we present \frameworkname, a contrastive
learning framework to address the script barrier in
the representation space of mPLMs
in a way that overcomes the limitations of prior work. To start with, a small portion of the data from 
the pretraining corpus of an mPLM 
is used to generate Latin transliteration, using \texttt{Uroman}\footnote{\url{https://github.com/isi-nlp/uroman}} \citep{hermjakob-etal-2018-box}. Then we create paired data using sentences in their original script and their transliterations. The data is subsequently used by the two objectives: Masked Language Modeling (MLM) and Transliteration Contrastive Modeling (TCM). MLM is applied to both the original sentences and their transliterations;
we use TCM to learn better-aligned representations by contrasting the positive pairs 
(paired data) against negative pairs (the remaining in-batch samples) as shown in Figure \ref{fig:model_illustration}.

Using Glot500-m \citep{imanigooghari-etal-2023-glot500} as
our source model, we evaluate \frameworkname both
``globally'' and ``locally''. Specifically, we fine-tune
Glot500-m on 5\% of its pretraining data of all languages
and refer to the resulting model as \modelname. We show
that \modelname aligns representations from different
scripts better and it generally outperforms the baselines on
sentence retrieval, sequence labeling, and text
classification tasks for different script groups. Our
ablation study indicates MLM and TCM in \frameworkname are
both important for achieving good crosslingual
performance. We additionally conduct a case study on Indic
languages, a group of languages that show areal features and
use different scripts. \modelnameIndi fine-tuned by \frameworkname using the data from Indic languages shows consistent improvement over the baseline.

The main contributions of this work are summarized as follows: 
(i) We present \frameworkname, a simple but effective framework, to address the script barrier in the representation space of mPLMs. (ii) We conduct extensive and controlled experiments on a variety of crosslingual tasks and show \frameworkname boosts performance. (iii) We show the framework encourages the representations from different scripts to be better aligned. (iv) In a case study on Indic languages, we demonstrate that \frameworkname also works for areal languages that have shared vocabulary but use distinct scripts.

\section{Related Work}
Transliteration refers to converting languages from one script into another script \citep{chou1981conversion}. Transliteration can increase lexical overlap, and therefore it has been shown to substantially improve the performance of neural machine translation for low-resource languages of different scripts \citep{gheini2019universal,goyal-etal-2020-efficient,amrhein-sennrich-2020-romanization}. Several studies also demonstrate that transliteration can enhance the crosslinguality of mPLMs
across various dimensions.
For instance, \citet{dhamecha-etal-2021-role} transliterate seven Indo-Aryan family languages into Devanagari  and show that common-script representations facilitate fine-tuning in a multilingual scenario. \citet{purkayastha-etal-2023-romanization} show that, by transliterating into Latin, better performance is achieved when adapting mPLMs to new languages, particularly for low-resource languages.
More recently, \citet{moosa-etal-2023-transliteration} focus on Indic languages and show that models directly pretrained on transliterated corpora in the Latin script achieve better performance.
However, the downside is that the model only supports one script and loses the ability to deal with the scripts in which the languages were originally written. This is not optimal when we expect predictions or generations in the original scripts.
In contrast to this line of work, we aim to directly break the script barrier, instead of circumventing it by limiting the model to one common script. We use transliterations in our fine-tuning framework to improve the alignment across different scripts: the model after fine-tuning still supports the scripts it originally did.

Contrastive learning is a method for learning meaningful
representations by contrasting positive pairs against
negative
pairs \citep{Chopra2005contrast,Hadsell2006contrast}. This
type of approach has achieved great success in learning
visual
representations \citep{schroff2015facenet,oord2018representation,chen2020contrastvisual,he2020momentum}. Contrastive
learning also demonstrates its effectiveness in NLP,
especially for learning sentence
representations \citep{gao-etal-2021-simcse,zhang-etal-2022-contrastive,wu-etal-2022-smoothed,zhang-etal-2023-contrastive-learning}. One
major problem in contrastive learning is how to construct
contrastive pairs. For a monolingual scenario, depending on
specific downstream tasks, the positive pairs are usually
constituted through data transformation or data augmentation
strategies \citep{zhang-etal-2020-unsupervised,yan-etal-2021-consert,wu-etal-2022-pcl,xu-etal-2023-simcse}
whereas the negative pairs are typically the remaining
in-batch samples \citep{Xu2023contrastivemethods}. In a
multilingual scenario where parallel data is available,
translations from different languages can be used to
construct the positive
pairs \citep{reimers-gurevych-2019-sentence,pan-etal-2021-contrastive,chi-etal-2021-infoxlm,wei2021universal}. Unfortunately,
large parallel corpora are mostly available for
high-resource languages. Therefore, aiming to improve
crosslinguality, especially for under-represented languages,
we start from the perspective of the script and construct
positive pairs by using sentences in their original script
and their Latin transliterations that can be easily obtained. Then we fine-tune an mPLM with our contrastive framework \frameworkname. In this way, our work also resembles some post-pretraining alignment approaches \citep{pan-etal-2021-multilingual,feng-etal-2022-language,ji-etal-2023-isotropic} that fine-tune a PLM using token-level or sentence-level translations.

\begin{figure*}
  \centering
  \includegraphics[width=\textwidth]{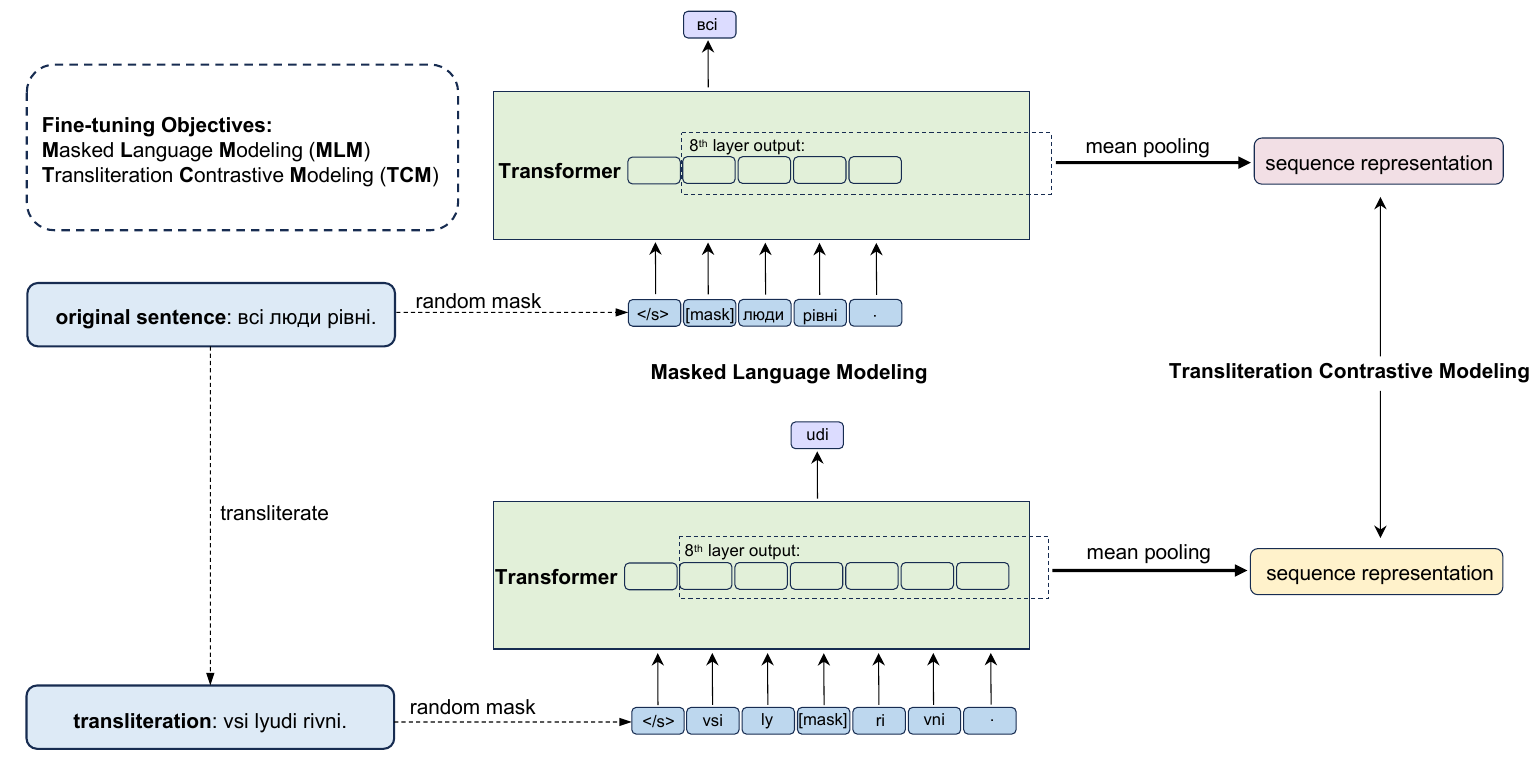}
  \caption{Overview of \textbf{\frameworkname}. We perform \textbf{Masked Language Modeling} for a sentence in its original script and its transliteration in the Latin script. Meanwhile, we calculate the sequence representations of the paired input by mean pooling their 8th layer output (ignoring the special token except for [mask] token). We then perform \textbf{Transliteration Contrastive Modeling} on the paired representations against negative pairs (not shown) in a batch.}
  \label{fig:framework}
\end{figure*}

\enote{hs}{the figure is great! it's a pity that you don't
show the MLM in complete analogy with the TCM (the reader
has to infer where MLM is used), but it's probalby ok}

\section{Methodology}
We present \textbf{\frameworkname}, a simple framework to address the script barrier by fine-tuning a PLM on a small portion of the data that is used to pretrain the model. The framework consists of two training objectives: Masked Language Modeling \citep{devlin-etal-2019-bert} and Transliteration Contrastive Modeling. We illustrate our framework in Figure \ref{fig:framework} and introduce our training objectives in the following.
\subsection{Masked Language Modeling}
The MLM training objective is to take an input sentence $X = [x_1, x_2, \cdots, x_n]$, randomly replace a certain percentage (15\% in our case) of tokens by [mask] tokens, and then train the model to predict the original tokens using an MLM head. Formally, let $\boldsymbol{H} = [\boldsymbol{h}_1, \boldsymbol{h}_2, \cdots, \boldsymbol{h}_n]$ be the contextualized representations at the last layer of the Transformer model \citep{vaswani2017transformer} (the output of the last Transformer block of the model) given the input sentence $X$. Following the notations used by \citet{meng2021cocolm}, we compute the MLM loss as follows:
\begin{equation*}
    \mathcal{L}_{\text{MLM}} = \mathbb{E}\left[- \sum_{i \in \mathcal{M}} \log p_{\text{MLM}}(x_i | \boldsymbol{h}_i)\right]
\end{equation*}
where $\mathcal{M}$ is the set of masked positions in the input sentence $X$ and $p_{\text{MLM}}(x_i | \boldsymbol{h}_i)$ is the probability of outputting the original token giving  $\boldsymbol{h}_i$ from the vocabulary $V$, computed by the MLM head. 

Instead of only performing MLM for the sentences in their original scripts, we also perform MLM for their transliterations in Latin script:  $X^{\text{trans}} = [x_1, x_2, \cdots, x_m]$, which are obtained by using \texttt{Uroman} \citep{hermjakob-etal-2018-box}. The MLM loss for transliteration data is referred to as
$\mathcal{L}_{\text{MLM}}^{\text{trans}}$.
By doing this, we can improve the crosslinguality of the
model across related languages that use different scripts,
as transliteration has shown to be effective in capturing
morphological
inflection \citep{murikinati-etal-2020-transliteration} and generating shared subwords between related languages \citep{muller-etal-2021-unseen,dhamecha-etal-2021-role,moosa-etal-2023-transliteration}. The intuition is that, as all sentences are consistently transliterated into Latin using the same tool, this can bring about vocabularies that have more shared subwords that are originally in different scripts. The improved lexical overlap therefore encourages the model to generate more crosslingual representations through MLM objective.

\enote{hs}{the above argument is not so clear. i guess the
assumption is that romanization generates shared subwords?
if there are not common subwords then it wouldn't work?}
\enote{yl}{yes the assumption you pointed out is correct. I slightly updated the paragraph above.}

\subsection{Transliteration Contrastive Modeling} 
Modeling a sentence and its transliterations separately does not necessarily lead to a good alignment between two types of scripts. Therefore, we propose to learn more similar and robust representations of a pair of sentences in different scripts using the contrastive learning objective. Sentence-level contrastive learning generally aims to align a positive pair of sentences by distinguishing them from negative or unrelated samples of sentences \citep{gao-etal-2021-simcse,meng2021cocolm,zhang-etal-2023-contrastive-learning}. In our framework, we simply let a positive pair be a sentence in its original script and its transliteration in the Latin script, and other sentences in a training batch be the negative samples to contrast with. 

Formally, let a training batch for TCM objective be $B
= \{X_1^{\text{orig}}, X_1^{\text{latn}}, \cdots,
X_N^{\text{orig}}, X_N^{\text{latn}}\}$, where $N$ is the
batch size, $X_i^{\text{orig}}$ is the $i$th sentence in its
original script and $X_i^{\text{latn}}$ is its Latin
transliteration. Similar to the setting
of \citet{meng2021cocolm}, a positive pair $(X, X^{+})$
consists of a symmetrical contrast pair, i.e., both
$(X_i^{\text{orig}}, X_i^{\text{latn}})$ and
$(X_i^{\text{latn}}, X_i^{\text{orig}})$, whereas the
negative samples are all the remaining sentences in their
original scripts and their transliterations in the training
batch using a slightly abusing notation:
$B^{-} = B  \setminus \{X, X^+\}$.
\enote{hs}{the above is an abuse of notation. is this the
orginal notation used by Meng et al? then it's probalby ok,
but you may want to add soemthing like ``slightly abusing notation'' } 
\enote{yl}{yes, this is the notation used by Meng et al. I added the phrase you suggested.}
Then the contrastive loss is defined as follows:
\begin{equation*}
    \mathcal{L}_{\text{TCM}} = \mathbb{E}\left[- \log \frac{\exp(\text{sim}(\boldsymbol{h}, \boldsymbol{h}^+)/\tau)}{\exp(\text{sim}(\boldsymbol{h}, \boldsymbol{h}^+)/\tau) + \text{NEG}}\right]
\end{equation*}
where $ \text{NEG} = \sum_{X^{-} \in B^{-}}\exp(\text{sim}(\boldsymbol{h}, \boldsymbol{h}^-)/\tau)$, $\text{sim}(\cdot)$ is the similarity measure (cosine similarity is used), $\tau$ is the temperature (set to 1 by default), $\boldsymbol{h}$, $\boldsymbol{h}^+$ and $\boldsymbol{h}^-$ are the representations of $X$, $X^+$ and $X^-$ respectively, which are computed by mean pooling the output of the 8th layer. Choosing the 8th layer is based on previous empirical findings, as the first layers are weak in terms of crosslinguality whereas the last layers are too specialized on the pretraining task \citep{jalili-sabet-etal-2020-simalign,chang-etal-2022-geometry}. The numerator improves the \emph{alignment}, i.e., encouraging the model to assign similar representations to similar samples, while the denominator improves the \emph{uniformity}, i.e., encouraging the representations to be uniformly distributed on the unit hypersphere \citep{wang2020contrastive}. 

Since all sentences are expected to have representations
similar to their Latin transliterations by the contrast training objective, this implicitly encourages sentences in different scripts to be in the same subspace. Intuitively, the Latin script acts as a bridge to connect all other scripts, and therefore all other scripts are better aligned. We show better script-neutrality is achieved in the representation space compared with the original mPLM in \secref{representation}.

\subsection{Overall Training}
The overall training objective of \frameworkname is then the sum of the MLM loss (from the original data and the transliteration data) and the TCM loss:
\begin{equation*}
    \mathcal{L}_{\text{training}} = \lambda_1 \mathcal{L}_{\text{MLM}} + \lambda_2 \mathcal{L}_{\text{MLM}}^{\text{trans}} + \lambda_3 \mathcal{L}_{\text{TCM}}
\end{equation*}
where $\lambda_1$, $\lambda_2$ and $\lambda_3$ are the weights for each loss. Following \citep{meng2021cocolm}, we set $\lambda_1 = \lambda_2 = \lambda_3 = 1$, as we also find the initial losses from each part during training are in similar magnitude. By fine-tuning an mPLM with this overall training objective, the model is expected to \textbf{(1)} not forget the language modeling ability gained in its pretraining phase; \textbf{(2)} be able to model sentences in both their original scripts and in their Latin transliterations and \textbf{(3)} learn to better align representations from different scripts in the same subspace.

\section{Experiments}
\subsection{Setups}
We use Glot500-m \citep{imanigooghari-etal-2023-glot500}, a
continued pretrained model from
XLM-R \citep{conneau-etal-2020-unsupervised}, as our source
mPLM. The training data of Glot500-m is Glot500-c, which
contains 1.5B sentences from 511 languages and 30 scripts
(534 language-scripts\footnote{A language-script is a
combination of the ISO 639-3 code and the script.} in
total). For each language-script, we randomly select 5\%
sentences from its training set in Glot500-c as the training
data. We then concatenate these sentences from all
language-scripts and
use \texttt{Uroman} \citep{hermjakob-etal-2018-box}, a tool
for universal romanization, to transliterate them into the Latin script. Finally, our training data consists of around 75M pairs (a pair is a sentence in the original script and its Latin transliteration). Examples of \texttt{Uroman} transliteration are shown in Table \ref{tab:transliteration}. Note that we also include the sentences originally in Latin script and perform transliteration for them. This is because we want to (1) preserve the model's ability to model languages in Latin script (2) increase lexical overlap by including data where diacritics are removed (done by \texttt{Uroman}) and (3) improve the overall robustness of the model. We show in \secref{ablation} how the model fine-tuned in this setting outperforms the model fine-tuned without Latin data. The fine-tuned model by using the proposed \frameworkname framework is referred to as \textbf{\modelname}. See \secref{hyperparam} for detailed hyperparameters. Except for the evaluation performed on the original datasets discussed in the main content below, we also evaluate the resulting models on the transliterated datasets. That is, we use \texttt{Uroman} to transliterate the datasets from the original script to the Latin script, and then perform evaluation on the new datasets. The performance on transliterated datasets is shown and discussed in \secref{trans_eval}.

\subsection{Downstream Tasks}

\paragraph{Sentence Retrieval.} Two datasets are considered: Tatoeba \citep{artetxe-schwenk-2019-massively} (SR-T) and Bible (SR-B). We select up to 1,000 English-aligned sentences for SR-T, following the same setting used by \citet{hu2020xtreme}; and up to 500 sentences for SR-B. We report the top-10 accuracy for both tasks. Following \citet{jalili-sabet-etal-2020-simalign}, the similarity is calculated by using the average of contextualized word embeddings at the 8th layer. 

\begin{table}[h]
    \setlength{\belowcaptionskip}{-0.3cm}
\centering
\scriptsize
\centering
\includegraphics[width=0.46\textwidth]{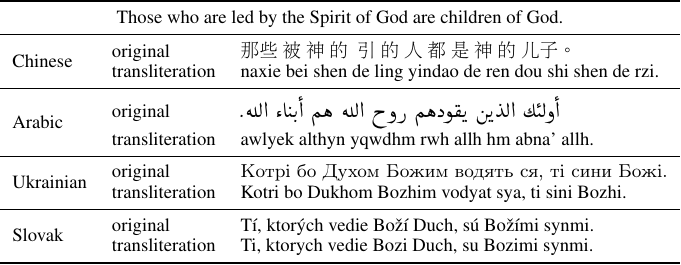}
\caption{Examples of \texttt{Uroman} transliteration. We select sentences (translations of the sentence ``Those who are led by the Spirit of God are children of God.'') from four languages that uses different scripts and transliterate them using \texttt{Uroman}. We notice some important characteristics of \texttt{Uroman}: tones (for Hani script) are not included and diacritics (for Latin script) are removed. }\label{tab:transliteration}
\end{table}

\paragraph{Text Classification.} Taxi1500 \citep{ma2023taxi1500}, a multilingual 6-class text classification dataset available in more than 1,500 languages, is used. We select a subset of language-scripts supported by the model for evaluation. We report the zero-shot crosslingual performance (in macro F1 scores) using the English train set for fine-tuning and selecting the best model on the English dev set.

\paragraph{Sequence Labeling.} Two types of tasks are considered: named entity recognition (NER) and Part-Of-Speech (POS) tagging. We use WikiANN  \citep{pan-etal-2017-cross} for NER  and Universal Dependencies \citep{de-marneffe-etal-2021-universal}, version v2.11, for POS. We report the zero-shot performance (in macro F1 scores) for both tasks.

\begin{table*}[h]
    \setlength{\belowcaptionskip}{-0.2cm}
    \scriptsize
    \centering
    \setlength{\tabcolsep}{1.0mm}{}
    \begin{tabular}{l|rrr|rrr|rrr|rrr|rrr}
        \toprule
        & \multicolumn{3}{c}{SR-B} & \multicolumn{3}{c}{SR-T} & \multicolumn{3}{c}{Taxi1500} & \multicolumn{3}{c}{NER} & \multicolumn{3}{c}{POS}\\
        \cmidrule(lr){2-4} \cmidrule(lr){5-7} \cmidrule(lr){8-10} \cmidrule(lr){11-13} \cmidrule(lr){14-16}
        & XLM-R & Glot500 & \modelname & XLM-R & Glot500 & \modelname & XLM-R & Glot500 & \modelname & XLM-R & Glot500 & \modelname & XLM-R & Glot500 & \modelname\\
        \midrule
        Latn & 16.2 & \underline{45.1} & \textbf{57.4} & 55.7 & \underline{69.1} & \textbf{73.0} & 22.5 & \underline{52.6} & \textbf{59.8} & 60.3 & \underline{66.1} & \textbf{67.3} & 68.1 & \underline{74.4} & \textbf{75.7} \\
        Cyrl & 25.5 & \underline{60.3} & \textbf{69.0} & 55.5 & \textbf{74.4} & \underline{69.7} & 30.2 & \underline{59.8} & \textbf{63.6} & 51.8 & \underline{65.3} & \textbf{66.2} & 66.7 & \underline{79.3} & \textbf{79.5} \\
        Hani & 30.4 & \textbf{43.4} & \underline{39.8} & \underline{62.0} & \textbf{80.5} & 47.7 & 66.6 & \underline{68.2} & \textbf{70.1} & \textbf{23.1} & \underline{22.2} & 21.9 & \underline{22.2} & \textbf{35.5} & 18.2 \\
        Arab & 36.3 & \underline{56.4} & \textbf{61.4} & 53.6 & \textbf{71.8} & \underline{56.3} & 48.5 & \underline{60.8} & \textbf{66.5} & 45.0 & \underline{53.4} & \textbf{57.7} & 65.8 & \underline{68.8} & \textbf{69.3} \\
        Deva & 32.1 & \underline{60.3} & \textbf{66.8} & 68.6 & \textbf{81.8} & \underline{71.9} & 49.5 & \underline{66.6} & \textbf{73.2} & \underline{56.9} & 56.2 & \textbf{58.9} & 58.3 & \underline{59.8} & \textbf{60.8} \\
        Other & 33.8 & \underline{49.0} & \textbf{53.6} & \underline{59.7} & \textbf{71.1} & 57.6 & 49.5 & \underline{59.5} & \textbf{65.2} & 45.2 & \textbf{50.4} & \textbf{50.4} & 65.9 & \textbf{68.8} & \underline{67.1} \\
        All & 19.3 & \underline{47.2} & \textbf{58.1} & 56.6 & \textbf{70.7} & \underline{68.8} & 26.7 & \underline{54.3} & \textbf{61.0} & 55.3 & \underline{61.6} & \textbf{62.8} & 65.6 & \underline{71.8} & \textbf{71.9} \\

        \bottomrule
    \end{tabular}
    \caption{Performance of \modelname and baselines on five downstream tasks across 5 seeds. We report the average performance for groups of languages using one of the five major scripts in the fine-tuning data: \textbf{Latn} (Latin), \textbf{Cyrl} (Cyrillic), \textbf{Hani} (Hani), \textbf{Arab} (Arabic), and \textbf{Deva} (Devanagari). We collect the remaining languages in the group ``\textbf{Other}''. In addition, we also report the average over all languages (group ``\textbf{All}''). \modelname generally performs better than other baselines except on SR-T.
    \textbf{Bold} (\underline{underlined}): best (second-best) result for each task in each group.}
    \label{tab:main_results}
\end{table*}

\begin{table*}[h]
    \setlength{\belowcaptionskip}{-0.4cm}
    \footnotesize
    \centering
    \setlength{\tabcolsep}{1.5mm}{}
    \begin{tabular}{lrrrrrrrrrrrrrrr}
        \toprule
        & \multicolumn{3}{c}{SR-B} & \multicolumn{3}{c}{SR-T} & \multicolumn{3}{c}{Taxi1500} & \multicolumn{3}{c}{NER} & \multicolumn{3}{c}{POS}\\
        \cmidrule(lr){2-4} \cmidrule(lr){5-7} \cmidrule(lr){8-10} \cmidrule(lr){11-13} \cmidrule(lr){14-16}
        & Latn & Other & All & Latn & Other & All & Latn & Other & All & Latn & Other & All & Latn & Other & All\\
        \midrule
        \textsc{Furina}$_{\text{-Latn}}$ & \textbf{52.1} & \textbf{58.6} & \textbf{53.5} & \textbf{65.9} & \underline{62.0} & \textbf{64.6} & \textbf{56.6} & \textbf{65.2} & \textbf{58.2} & 66.1 & \underline{53.6} & \underline{61.5} & \textbf{73.4} & \underline{66.7} & \textbf{70.9}  \\
        - w/o TCM & \underline{41.2} & \underline{58.5} & \underline{44.9} & \underline{57.6} & \textbf{67.8} & \underline{61.2} & \underline{52.9} & \underline{63.4} & \underline{54.9} & \textbf{66.2} & \textbf{54.4} & \textbf{61.9} & \underline{72.7} & 65.5 & 70.0 \\
        - w/o MLM & 31.2 & 26.3 & 30.2 & 40.2 & 25.4 & 35.1 & 44.8 & 52.8 & 46.4 & \underline{66.4} & 48.3 & 59.8 & 72.6 & \textbf{66.8} & \underline{70.4}  \\
        \midrule
        \midrule
        \textsc{Furina}$_{\text{+Latn}}$ & \textbf{57.4} & \textbf{60.8} & \textbf{58.1} & \textbf{73.0} & \underline{60.9} & \textbf{68.8} & \textbf{59.8} & \textbf{65.8} & \textbf{61.0} & \textbf{67.3} & \textbf{54.9} & \textbf{62.8} & \underline{75.7} & \underline{65.5} & \textbf{71.9}  \\
        - w/o TCM & \underline{45.5} & \underline{58.4} & \underline{48.3} & \underline{65.9} & \textbf{67.1} & \underline{66.3} & \underline{57.8} & \underline{64.4} & \underline{59.1} & \underline{67.2} & \underline{54.2} & \underline{62.4} & \textbf{75.9} & 64.5 & \underline{71.6}  \\
        - w/o MLM & 41.2 & 35.9 & 40.0 & 63.4 & 41.6 & 55.9 & 50.8 & 55.2 & 51.7 & 66.6 & 47.2 & 59.5 & 73.9 & \textbf{66.5} & 71.1  \\
        \bottomrule
    \end{tabular}
    \caption{Ablation study. We investigate the effect of incorporating Latin script data and their \texttt{Uroman} transliteration (models are therefore classified into two groups: \textsc{Furina}$_{\text{-Latn}}$ and \textsc{Furina}$_{\text{+Latn}}$). We also explore the influence of the MLM and TCM objectives. The model generally performs worse when any one of the objectives is missing. In addition, including Latin script data can improve the overall performance for both Latin script languages and languages using other scripts on all tasks.
    \textbf{Bold} (\underline{underlined}): best (second-best) result per controlled group.}
    \label{tab:ablation}
\end{table*}

\subsection{Results and Discussion}
To better illustrate how the proposed \frameworkname framework can influence the performance of different scripts, we group all language-scripts by their scripts and report the average performance for each group. The results of XLM-R, Glot500-m, and \modelname are shown in Table \ref{tab:main_results}. We see an overall improvement for \modelname compared with Glot500-m in each task except for SR-T. We conjecture that the sub-optimal performance on SR-T can be related to the domain shift and the small set of languages supported by Tatoeba. Our fine-tuning data consists of sentences of many low-resource languages that come from genre-specific corpora such as the Bible, which can be quite different from Tatoeba, which is more modern in terms of the genre and mostly only supports high-resource languages (70 out 98 languages are high-resource languages).

\modelname performs surprisingly well on ST-B. An overall
    improvement of 10.9 (from 47.2 to 58.1) is achieved. We
    also see a consistently large increase for each script
    group of languages: e.g., 12.3 for Latin script
    languages and 8.7 for Cyrillic languages. However, for
    Hani script (Chinese characters) languages, we see a sudden
    drop in performance, which can also be seen in other
    tasks such as SR-T, NER, and POS. 
    We hypothesize that \texttt{Uroman} is suboptimal for Hani script because a great deal of important information is lost in the transliteration: both due to the removal of tones and due to the conflation of semantically different characters that are pronounced identically (e.g., 
    \begin{CJK}{UTF8}{bsmi}氦\end{CJK} (helium) v.s  \begin{CJK}{UTF8}{bsmi}害\end{CJK} (harm), both pronounced hài in Mandarin). In addition, Chinese characters are logograms. Even if the transliteration contains correct tones, the transliterated words potentially lose semantic or contextual nuances and are more prone to ambiguity \citep{amrhein-sennrich-2020-romanization}, thus resulting in a performance drop.
However, note that Hani languages are high-resource languages, so this
result does not diminish our hypothesis that
transliteration helps low-resource languages.

For other types of tasks, we also see a consistent
improvement. In both NER and POS, \modelname achieves better
performance than Glot500-m in 4 out of 5 script groups
(except for Hani as discussed above). Compared with
token-level classification (NER and POS), we see a more
prominent increase in sequence-level classification
(Taxi1500): the overall F1 score is increased by 6.7 (more
than 10\%) compared with Glot500-m, and \modelname achieves
substantially better performance for each script
group.
The results indicate
that the proposed contrastive learning framework \frameworkname
boosts  crosslingual transfer learning,
especially for sequence-level tasks, e.g., sentence
retrieval and sentence classification. We also evaluate both Glot500m and FURINA under the
common script scenario, i.e., transliterating the evaluation dataset to Latin script. See the evaluation in \secref{trans_eval}.

\begin{figure*}
    \setlength{\belowcaptionskip}{-0.4cm}
\begin{tabular}{c}
\hspace{-0.3cm}
\includegraphics[width=0.24\textwidth]{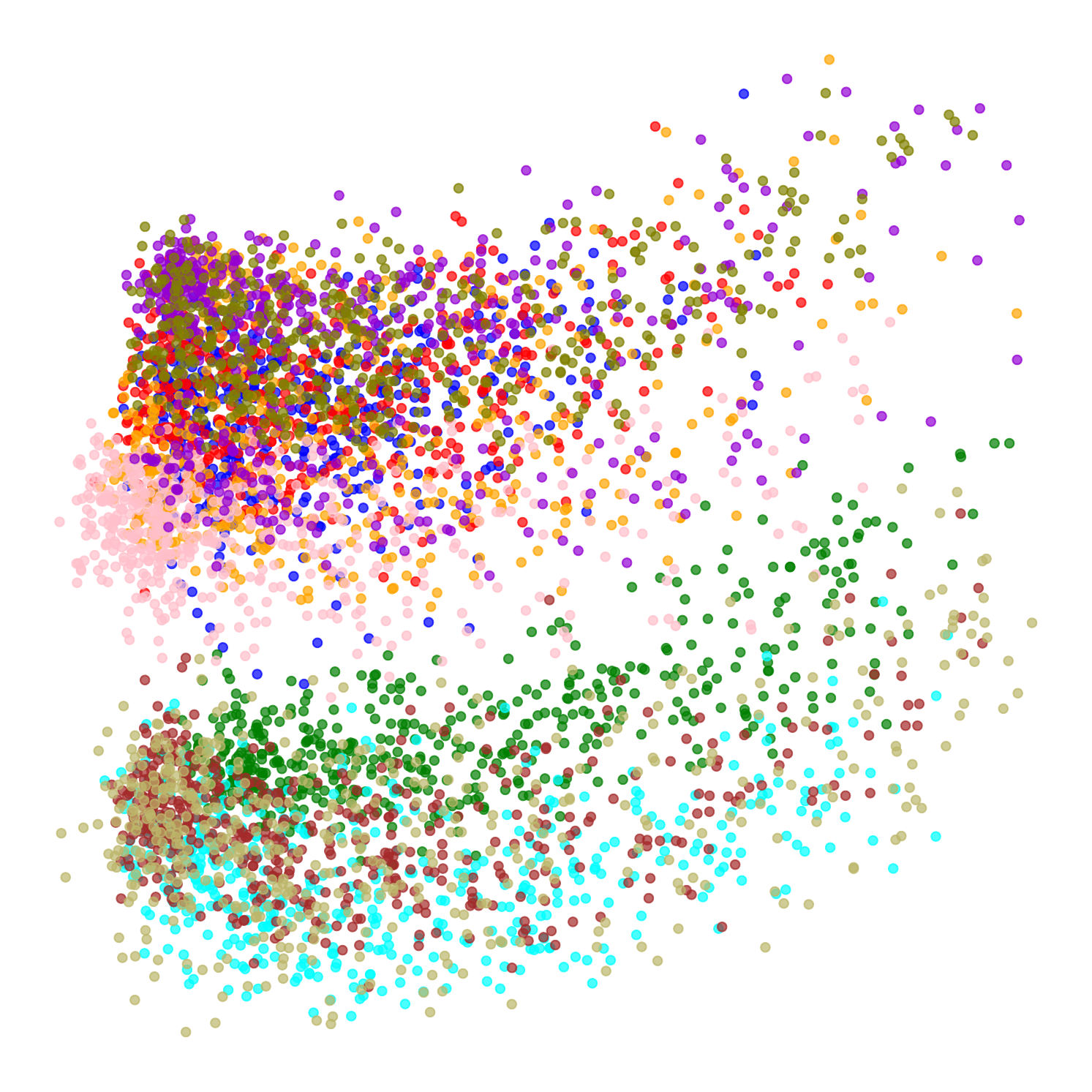}
\hspace{-0.3cm}
\includegraphics[width=0.24\textwidth,height=4cm]{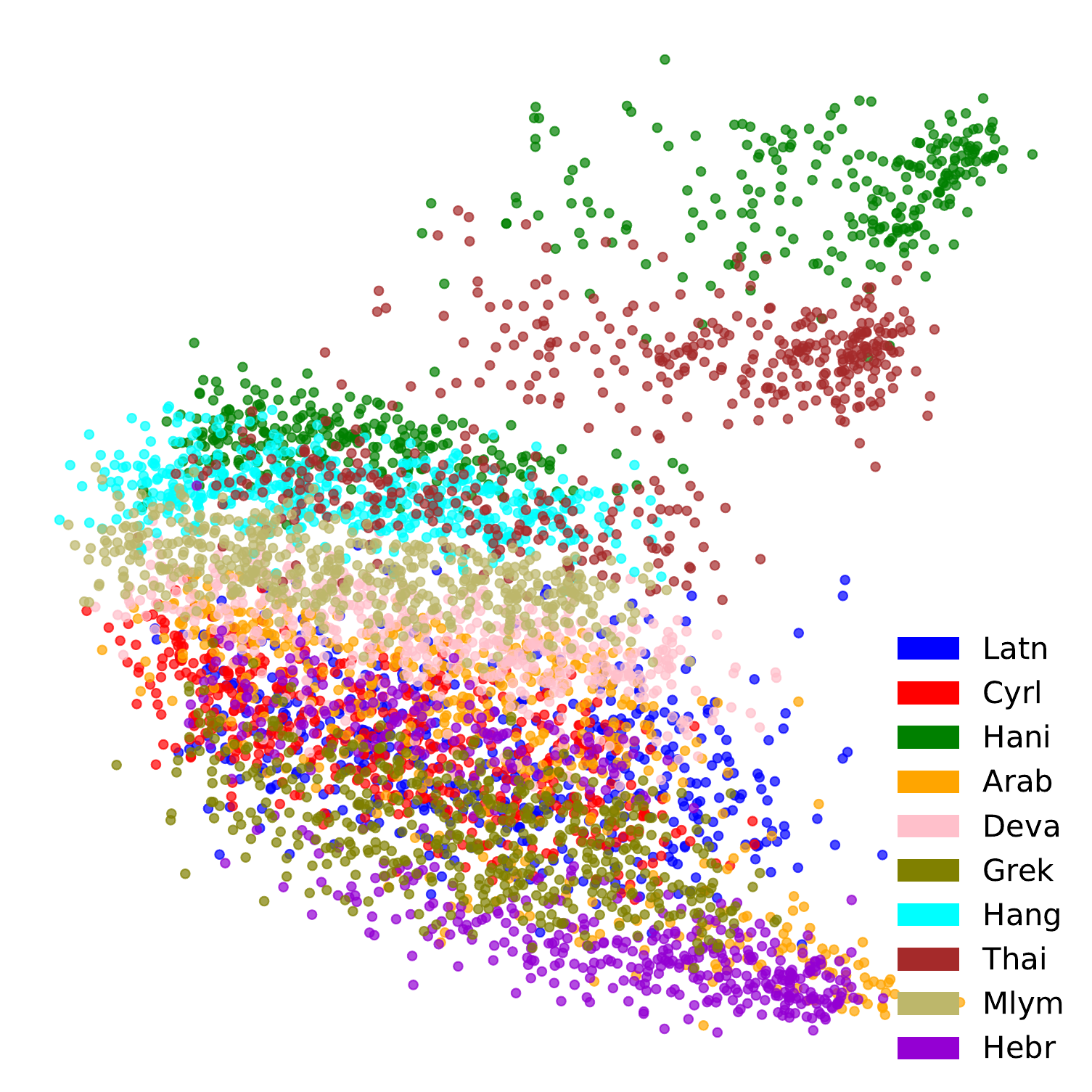}
\hspace{0.4cm}
\includegraphics[width=0.22\textwidth,height=4cm]{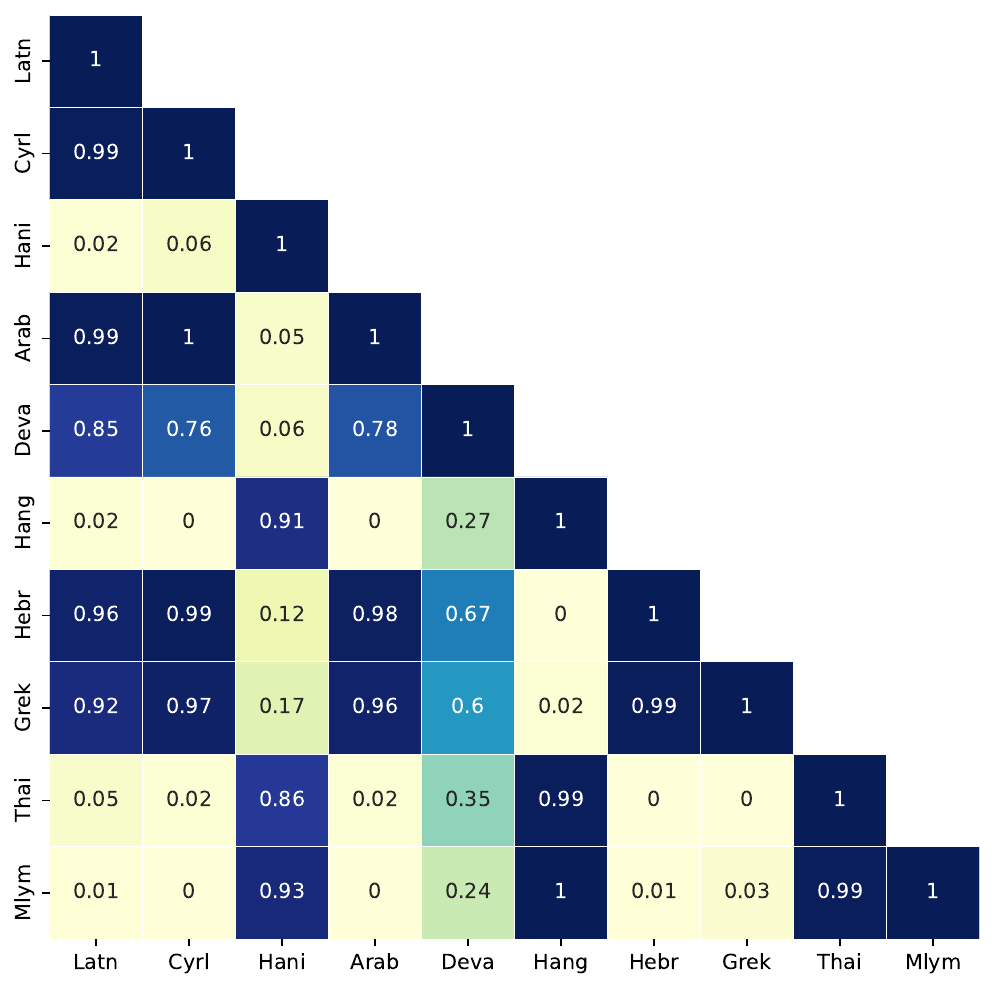}
\hspace{0.1cm}
\includegraphics[width=0.25\textwidth,height=4.05cm]{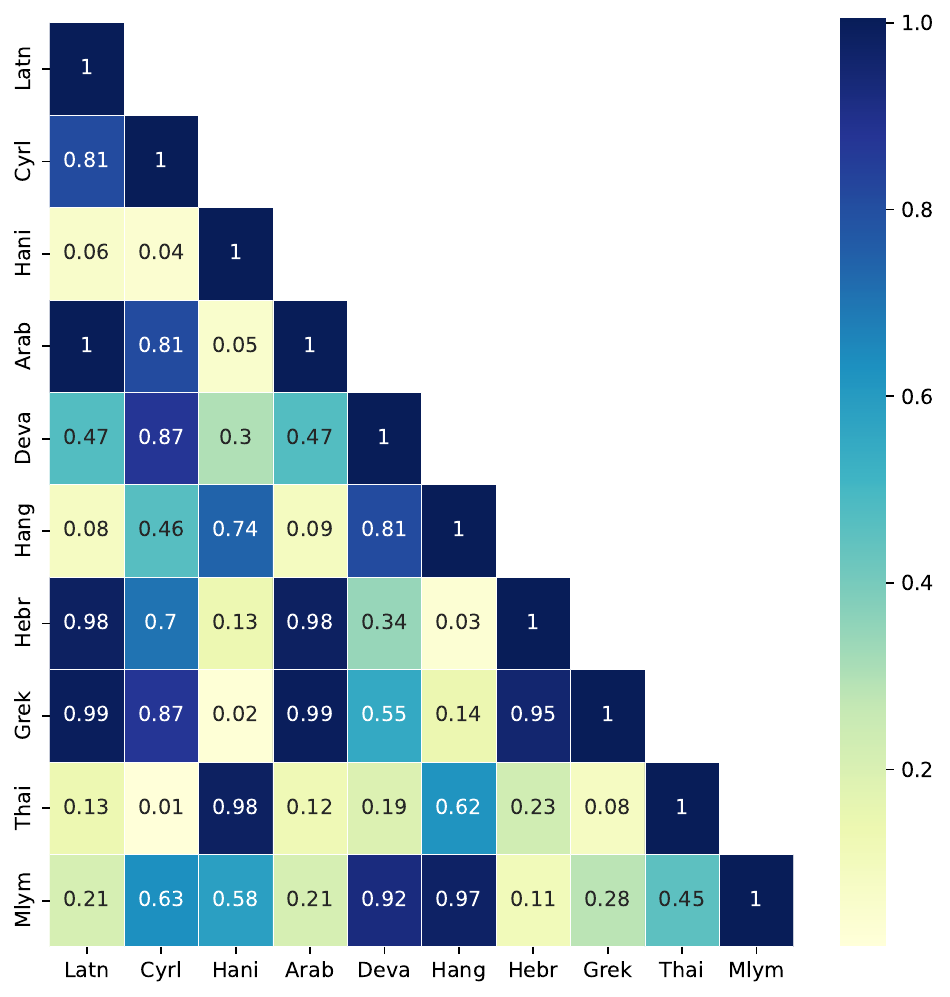}
\hspace{0.1cm}
\end{tabular}

\caption{(i)
PCA of sentence representations from 
layer 8 (mean-pooling of contextualized token embeddings,
dim=768)
of \modelname (Subfigure 1) and Glot500-m (Subfigure 2).
Points
are sentence representations. Colors
indicate scripts.
(ii) Pairwise cosine similarity
between centroids of scripts 
of \modelname (Subfigure 3) and Glot500-m (Subfigure 4).
\modelname  better
represents scripts in several cases, e.g., 
it better aligns related scripts \textcolor{Latn_color}{Latn}
and \textcolor{Cyrl_color}{Cyrl}, and,
it better
separates the unrelated scripts \textcolor{Cyrl_color}{Cyrl}
and \textcolor{Mlym_color}{Mlym}
compared to Glot500-m.}
\label{fig: PCA_layer_eight}
\end{figure*}

\section{Analysis}\seclabel{analysis}
\subsection{Ablation Study}\seclabel{ablation}
We conduct ablation experiments on all five tasks, using the same hyperparameters and Glot500-m as the source model. Specifically, we explore the influence of MLM and TCM objectives in \frameworkname. In addition, we investigate the importance of incorporating data from Latin script languages and classify the model variants into two groups (with Latin script languages or without). In our preliminary experiments, we also explore the weight of the TCM loss (e.g., 0.1, 0.5, and 1) and find the influence is very small as the TCM loss quickly goes to a small value for different weights, possibly due to the simplicity of the contrastive task (the initial magnitude is close though).
We report the best result for each model variant in Table \ref{tab:ablation}.

The model generally performs worse when any one of the
training objectives is missing. This holds for each
group. For example, the overall accuracy in SR-B decreases
by 8.6 and 13.3 for \textsc{Furina}$_{\text{-Latn}}$  when
TCM or MLM is missing. The only exception is the accuracy
for
``Other'' in SR-T. We notice when TCM is missing, the result
increases by 5.8
(\textsc{Furina}$_{\text{-Latn}}$)
and 6.2 (\textsc{Furina}$_{\text{+Latn}}$). The possible
reason is as follows. The sentences from languages using
different scripts in SR-T (Tatoeba sentences are quite simple) are already aligned pretty well. Additionally fine-tuning the model with TCM on a small portion of data (for many low-resource languages the data is related to the Bible) whose domain is different from the domain of SR-T hurts the performance of underrepresented languages. 

Interestingly, we find the introduction of  TCM and MLM
objectives has a more prominent impact on sequence-level
(SR-B, SR-T, Taxi1500) than  token-level tasks (NER,
POS). For example, for \textsc{Furina}$_{\text{-Latn}}$, all
three variants achieve competitive overall performance
on token-level tasks:
around 60 and 70 F1 scores on NER and POS respectively. This
might suggest that \frameworkname has a relatively small
effect on individual token representations as we use
sentence-level contrastive learning. In addition, in sequence
labeling, the model may be able to
transfer prevalent classes such as \textit{verb}
and \textit{noun} \citep{imanigooghari-etal-2023-glot500,liu2023ofa}
to some extent even without the explicit crosslingual
constraint imposed by TCM and MLM.


\enote{hs}{as a reviewer i would not see a big differnece
    between the two graphs. can you compute some objective
    measures of overlap / non-overlap? for example, the
    median cosine between the script groups?}
\enote{cl}{Yes, that make sense, I add pair-wise normalized cosine similarity
to make our argument more convicing.}

We also see a consistent improvement when the Latin script data is incorporated into the fine-tuning data. Although there is an occasional small decrease for languages using other scripts (e.g., on SR-T and POS) when comparing \textsc{Furina}$_{\text{+Latn}}$ and \textsc{Furina}$_{\text{-Latn}}$, the increase for Latin script languages is prominent for each task. This is expected, as \textsc{Furina}$_{\text{-Latn}}$ can catastrophically forget the knowledge gained in its pretraining phase \citep{Kirkpatrick2017catastrophic}. By incorporating the Latin script data and their \texttt{Uroman} transliteration, we can further increase lexical overlap 
and make the model more robust, since \texttt{Uroman} has a unified mechanism for romanization and removes all diacritics. For example, the word ``sal\'on'' in Czech will be the same as the French word ``salon'' after \texttt{Uroman} removing the diacritics on ``\'o''.

\subsection{Representation Visualization}\seclabel{representation}

\begin{table*}[hbt!]
\setlength{\belowcaptionskip}{-0.6cm}
\begin{center}
\begin{adjustbox}{max width=\textwidth}
\small
\setlength{\tabcolsep}{3mm}{}
\begin{tabular}{l c c c c c c c c c c c c}
\toprule[0.8pt]
\textbf{Language} & \textbf{pan} & \textbf{hin} & \textbf{ben} & \textbf{ori} & \textbf{asm} & \textbf{guj} & \textbf{mar} & \textbf{kan} & \textbf{tel} & \textbf{mal} & \textbf{tam} & \textbf{avg} \\
\midrule[0.8pt]
\multicolumn{13}{l}{\textbf{Named Entity Recognition (F1-Score)}}\\
    \modelnameIndi & 92.5&\textbf{97.1}&{98.6}&96.5&\textbf{94.5}&\textbf{95.7}&\textbf{97.2}&\textbf{93.9}&\textbf{96.7}&\textbf{97.2}&\textbf{97.0}&\textbf{96.1}\\
Glot500-m   & 
\textbf{92.7}&\textbf{97.1}&\textbf{98.7}&\textbf{96.6}&93.5&95.4&97.1&93.5&\textbf{96.7}&97.0&\textbf{97.0}&96.0\\
    \textcolor{gray}{ALBERT\textsubscript{$US$}}   & \textcolor{gray}{85.4} &\textcolor{gray}{92.9} & \textcolor{gray}{97.3} & \textcolor{gray}{93.5} & \textcolor{gray}{89.1} & \textcolor{gray}{80.2} & \textcolor{gray}{90.6} & \textcolor{gray}{-} &\textcolor{gray}{-} & \textcolor{gray}{-} & \textcolor{gray}{-} & \textcolor{gray}{89.9}\\
\midrule[0.5pt]
\midrule[0.5pt]
\multicolumn{13}{l}{\textbf{\textbf{Wikipedia Section Title Prediction (Accuracy)}}}\\
\modelnameIndi  & 81.2&\textbf{83.9}&\textbf{85.8}&\textbf{83.7}&\textbf{85.2}&\textbf{84.8}&\textbf{85.6}&\textbf{84.3}&\textbf{96.6}&\textbf{83.8}&\textbf{83.8}&\textbf{85.3}\\
Glot500-m  & \textbf{81.5}&83.6&85.5&83.2&\textbf{85.2}&84.4&84.7&83.9&\textbf{96.6}&83.1&83.6&85.0\\
\textcolor{gray}{ALBERT\textsubscript{$US$}}   & \textcolor{gray}{77.6} & \textcolor{gray}{82.2} & \textcolor{gray}{84.4} & \textcolor{gray}{81.5} & \textcolor{gray}{81.7} & \textcolor{gray}{82.4} & \textcolor{gray}{82.7} & \textcolor{gray}{-} &\textcolor{gray}{-} & \textcolor{gray}{-} & \textcolor{gray}{-} & \textcolor{gray}{81.8} \\
\midrule[0.5pt]
\midrule[0.5pt]
\multicolumn{13}{l}{\textbf{\textbf{Cloze Style Question Answering (Accuracy)}}}\\
\modelnameIndi  & \textbf{40.6}&\textbf{46.8}&\textbf{44.9}&\textbf{45.2}&\textbf{47.3}&\textbf{85.9}&\textbf{51.7}&\textbf{39.7}&\textbf{32.3}&\textbf{38.2}&\textbf{37.7}&\textbf{46.4}\\
Glot500-m  & 29.8&28.8&27.7&29.4&31.8&82.7&32.1&28.9&26.3&27.5&29.3&34.0\\
\textcolor{gray}{ALBERT\textsubscript{$US$}}  &  \textcolor{gray}{32.8}& \textcolor{gray}{38.5}& \textcolor{gray}{36.4}& \textcolor{gray}{36.0}& \textcolor{gray}{37.4}& \textcolor{gray}{70.2}& \textcolor{gray}{39.5}& \textcolor{gray}{-}&\textcolor{gray}{-} &\textcolor{gray}{-} & \textcolor{gray}{-} & \textcolor{gray}{41.5}  \\
\bottomrule[0.8pt]
\end{tabular}%
\end{adjustbox}
\caption{Evaluation of NER, WSTP, and CSQA (zero-shot) from IndicGLUE Benchmark. \modelnameIndi consistently outperforms Glot500-m in most languages on three tasks. \textbf{Bold}: best result for each language in each task. We also show \textcolor{gray}{ALBERT\textsubscript{$US$}} model trained by \citet{moosa-etal-2023-transliteration} using only romanized data for reference. \textcolor{gray}{ALBERT\textsubscript{$US$}} cannot be compared with the other two models directly, because it uses different pretraining data.
}
\label{table-res}
\end{center}
\end{table*}

To explore how \frameworkname manipulates the representation space, we visualize the sentence representations from languages that use different scripts. Specifically, we feed Glot500-m and \modelname with 500 sentences from the SR-B task. To facilitate comparison, we only select 10 high-resource languages. Each language uses one of the 10 dominant scripts in the vocabulary of the models (we use GlotScript \citep{kargaran-etal-2024-glotscript-resource} to detect the script of the tokens). The languages are English (\textcolor{Latn_color}{Latin}), Russian ( \textcolor{Cyrl_color}{Cyrl}), Chinese (\textcolor{Hani_color}{Hani}), Arabic (\textcolor{Arab_color}{Arab}), Hindi (\textcolor{Deva_color}{Deva}), Greek (\textcolor{Grek_color}{Grek}), Korean (\textcolor{Hang_color}{Hang}), Malayalam (\textcolor{Mlym_color}{Mlym}), and Hebrew (\textcolor{Hebrew_color}{Hebrew}).
We obtain the sentence representations by mean-pooling the
contextualized token embeddings. We visualize the
representations of the 8th layer (also used for SR-B and SR-T) by
projecting them to two dimensions with principal component analysis (PCA) in Figure \ref{fig: PCA_layer_eight} (1st and 2nd subfigures). We also compute the pairwise normalized cosine similarity between the centroid of each script group (3rd and 4th subfigures). Additionally, we visualize representations from every layer in Appendix \ref{appendix:visual_appendix}.

It can be seen that the representations from each script
roughly form an individual single cluster for Glot500-m. 
In contrast, representations only form two major clusters
for \modelname, where each cluster has 
certain related scripts.
This is evidence that Glot500-m encodes
script-sensitive information and distinct but related
scripts are not well-aligned,
whereas \modelname has learned better
script-neutrality for the representations 
of related scripts. 
The pairwise similarity also supports our argument:
e.g., \modelname has higher similarity scores among (1) \textcolor{Latn_color}{Latn}, \textcolor{Cyrl_color}{Cyrl}, and \textcolor{Deva_color}{Deva}, and (2) \textcolor{Hani_color}{Hani}, \textcolor{Thai_color}{Thai}
and \textcolor{Hang_color}{Hang}, where any script in each group is related to the rest of the scripts.
This phenomenon
indicates that \frameworkname is effective in addressing the
script barrier in the representation space of
mPLMs by improving the similarity of the related scripts.
Interestingly, we observe linguistic features and
geographical proximity can also be related to the
representation
subspaces. \textcolor{Hani_color}{Hani}, \textcolor{Thai_color}{Thai}
and \textcolor{Hang_color}{Hang} (Hangul) are in the same cluster, which might be explained by the fact that Thai, Korean, and Chinese are spoken in adjacent areas and their vocabularies are mutually influenced. With \frameworkname, such similarity is further exploited and thus their representations are encouraged to be closer.

\begin{table}[hbt!]
    \setlength{\abovecaptionskip}{0.2cm}
    \setlength{\belowcaptionskip}{-0.3cm}
\begin{center}
\begin{adjustbox}{max width=.45\textwidth}
\small
\begin{tabular}{l r r r r}
\\ \hline 
\textbf{Lang.} & \textbf{Sub-family} & \textbf{Script}
    & \textbf{\# Sentence}\\
\hline 
asm & Eastern Indo-Aryan & Bengali & 188.0k\\
bhi & Eastern Indo-Aryan & Devanagari & 4351.4k\\
guj & Western Indo-Aryan & Gujarati & 3.0k\\
guj & Western Indo-Aryan & Devanagari & 4.7k \\
mai & Eastern Indo-Aryan & Devanagari & 4573.8k\\
nep & Northern Indo-Aryan & Devanagari & 704.5k\\
pan & Northwestern Indo-Aryan & Gurmukhi & 5.3k\\
sin & Insular Indo-Aryan & Sinhala & 2874.8k\\
ben & Eastern Indo-Aryan & Bengali & 131.5k\\
hin & Central Indo-Aryan & Devanagari & 40.9k\\
mar & Southern Indo-Aryan & Devanagari & 2905.1k\\
ori & Eastern Indo-Aryan & Oriya & 16.4k\\
sam & Sanskrit & Devanagari & 729.2k\\
\hline 
\end{tabular}
\end{adjustbox}
\caption{The 12 languages in the Indic group used for fine-tuning \modelnameIndi. The number of sentences shown is the result of randomly sampling 10\% of sentences from Glot500-c 
for each language.}
\label{table-indic-group}
\end{center}
\end{table}

\subsection{Case Study: Indic Group}\seclabel{indic}
To further explore  \frameworkname's performance, we conduct
a case study on 12 languages that exhibit areal features such as shared vocabulary. 
These languages are mostly from the Indo-Aryan group
and are mutually influenced with each other linguistically, historically,
and phonologically, but use different scripts \citep{moosa-etal-2023-transliteration}. 

Similar to the previous settings, we use Glot500-m as our
source mPLM with the training data as the only difference.
Specifically, we randomly sample 10\% of sentences from
Glot500-c \citep{imanigooghari-etal-2023-glot500} for each
of the 12 languages. The details of the resulting
fine-tuning dataset are shown in
Table \ref{table-indic-group}. We then use \texttt{Uroman}
to transliterate these sentences into the Latin
script. Subsequently, we create positive paired data using
these sentences and their Latin script
transliterations. This results in our training data
consisting of 16.5M sentence 
pairs. We use \frameworkname to fine-tune Glot500-m on the
data and refer to the final model
as \modelnameIndi. Following the similar setting employed
by \citet{moosa-etal-2023-transliteration}, we
evaluate \modelnameIndi with three downstream tasks from
IndicGLUE \citep{kakwani-etal-2020-indicnlpsuite}:
Wikipedia Section
Title Prediction (WSTP),  Cloze Style Question
Answering (CSQA)) and a Named Entity Recognition (NER)
task. One major difference between our evaluation
and \citet{moosa-etal-2023-transliteration} is that we
always evaluate the languages in their \textbf{original}
scripts, whereas \citet{moosa-etal-2023-transliteration}
evaluate on the transliterated data in a unified script
(Latin). The details for hyperparameters of each task are
reported in \S\ref{appendix:indic_appendix}. To
test the crosslingual transfer ability of \modelnameIndi on
the Indic group languages, for WSTP and NER, we fine-tune
the model on all languages at once and then evaluate the
model on the test set for each language. We use CSQA  to test the
zero-shot capability of \modelnameIndi. We evaluate on F1
for NER and on accuracy for WSTP and CSQA. The results are
shown in Table \ref{table-res}.

\modelnameIndi outperforms Glot500-m on 8 out of 11 languages on the NER task, with a 0.1 slightly higher average score than Glot500-m. 
We see a similar small improvement on WSTP,
where \modelnameIndi beats Glot500-m on most of the
languages except for Panjabi. We assume the reason for the
small improvement in these two tasks is that the model is
fine-tuned on the train set of \textbf{all} languages (not
zero-shot crosslingual transfer), which could overshadow the
benefits from \frameworkname. Nevertheless, the general
consistent improvement shows the effectiveness
of \frameworkname.
For the zero-shot task CSQA,
we see a large increase. \modelnameIndi improves the average
performance by more than 12 points (from 34.0 to
46.4). Although another competitive model, \textcolor{gray}
{ALBERT\textsubscript{$US$}} \citep{moosa-etal-2023-transliteration},
beats Glot500-m
(source model of \modelnameIndi), 
\modelnameIndi
outperforms \textcolor{gray}{ALBERT\textsubscript{$US$}}
consistently in all languages. This indicates
that \frameworkname enjoys the largest improvements in
zero-shot scenarios, which is exactly the
desideratum for most low-resource languages. Overall, the
consistent improvement demonstrates \frameworkname not only works well ``globally'' on all languages but also ``locally'' on a small group of languages that have lexical overlap but use different scripts.

\section{Conclusion}
In this work, we propose a novel framework \frameworkname to fine-tune an mPLM only using a small portion of sentences in its pretraining corpus and their Latin transliteration. The framework contrasts the sentences in their original script and their transliterations to tackle the script barrier problem. Using Glot500-m as our source model, we fine-tune it using the proposed framework and present the resulting model: \modelname. Through extensive experiments, we show \modelname better aligns the representations from different scripts into a common space and therefore outperforms Glot500-m on a wide range of crosslingual transfer tasks. In addition, we conduct a case study on Indic group languages that are known to be mutually influenced by each other but use different scripts. We show \frameworkname can also boost the performance for the Indic group languages. We hope this framework can inspire more future work leveraging transliteration to improve crosslinguality of mPLMs.

\section*{Limitations}
We propose a simple contrastive learning framework \frameworkname that aims to address the script barrier. We show the effectiveness by using Glot500-m as the source model and fine-tuning it on a small portion (5\%) of its pretraining data. We would assume the proposed framework can also be used directly for pretraining, through which the model might benefit further from seeing more data. However, due to a limited computation budget, we aren't able to pretrain a model from scratch or continued pretrain a model using the full Glot500-c corpus. We would leave out how the proposed framework can be integrated into efficient pretraining or continued pretraining for future work.

We see the proposed framework \frameworkname works ``globally'' when fine-tuning the model on data of all languages scripts, and also ``locally'' for a case study when only fine-tuning on the Indic-group languages that are mutually influenced and have extensive lexical overlap. Unfortunately, we didn't validate the framework further by trying more language groups, which could further demonstrate the usage of  \frameworkname. However, this is beyond the scope of this paper. Nevertheless, we hope our framework can inspire more work that applies a similar framework and focus ``locally'' on groups of languages of interest for future research.

\section*{Acknowledgements}
This work was funded by the European Research
Council (NonSequeToR, grant \#740516). We appreciate Ayyoob
Imani’s suggestions for designing a case study
and Lixi Liu's suggestions for the pairwise 
similarity graph. We also thank the anonymous reviewers for their constructive feedback.

\bibliography{anthology,custom}

\appendix

\section{Hyperparameters}\seclabel{hyperparam}

\subsection{Fine-tuning on Glot500-c}\label{Fine-tuning_on_Glot500-c}

\begin{table*}
  \footnotesize
	\centering
	\def\tablesep{0.1cm}
\begin{tabular}{
  @{\hspace{\tablesep}}l@{\hspace{\tablesep}}|
  @{\hspace{\tablesep}}r@{\hspace{\tablesep}}
  @{\hspace{\tablesep}}r@{\hspace{\tablesep}}
  @{\hspace{\tablesep}}r@{\hspace{\tablesep}}
  @{\hspace{\tablesep}}r@{\hspace{\tablesep}}
  @{\hspace{\tablesep}}r@{\hspace{\tablesep}}
  @{\hspace{\tablesep}}r@{\hspace{\tablesep}}
  @{\hspace{\tablesep}}r@{\hspace{\tablesep}}
  @{\hspace{\tablesep}}r@{\hspace{\tablesep}}
  @{\hspace{\tablesep}}r@{\hspace{\tablesep}}
}
 & indo1319 &atla1278 &aust1307 &turk1311 &sino1245 &maya1287 &afro1255 &other &all \\
\midrule
SR-B &93 &69 &55 &23 &23 &15 &12 &79 &369 \\
SR-T &54 &2 &7 &7 &3 &0 &5 &20 &98 \\
Taxi1500 &87 &68 &51 &18 &22 &15 &11 &79 &351 \\
NER &94 &5 &12 &12 &7 &0 &6 &28 &164 \\
POS &54 &2 &4 &5 &3 &1 &6 &16 &91 \\
  \end{tabular}
  \caption{The number of languages in each language family in downstream tasks.}
  \label{evaluation_info_family}
\end{table*}

We fine-tune Glot500-m \citep{imanigooghari-etal-2023-glot500} on a small portion of its training data, Glot500-c. Specifically, from each language-script, we randomly select 5\% sentences. Note that we also consider languages that use Latin scripts when constructing our training data (another variant is only considering languages that do not use Latin scripts, which we do for the ablation study described in \secref{ablation}). Then we construct paired data by generating the Latin transliterations for each sentence using \texttt{Uroman} \citep{hermjakob-etal-2018-box}. The paired data is then used to fine-tune the model using the proposed contrastive learning framework. For the MLM objective, we use the standard mask rate of 15\% for both sentences in their original scripts and their \texttt{Uroman} transliterations. The weights for all objectives are set to 1 by default. We use Adam optimizer \citep{ba2015adam} with $(\beta_1, \beta_2) = (0.9, 0.999)$ and $\epsilon = \text{1e-6}$. The initial learning rate is set to 1e-5. The effective batch size is set to 768. Each batch contains sentence pairs (one in its original script and one in Latin transliteration) randomly picked from all languages. We set the per-GPU batch to 24, the gradient accumulation to 8, and train on four RTX A6000 GPUs ( $ 24 \times 8 \times 4 = 768$). We use FP16 training (mixed precision \citep{paulius2018mixed}) by default. We store checkpoints for the model every 2K steps and apply early
stopping with the best average performance on
downstream tasks. The model is fine-tuned for a maximum
of 3 days (roughly 1 epoch).

\subsection{Fine-tuning on Downstream tasks}

\begin{table}[t]
  \small
	\centering
	\def\tablesep{0.2cm}
\begin{tabular}{
  @{\hspace{\tablesep}}l@{\hspace{\tablesep}}|
  @{\hspace{\tablesep}}r@{\hspace{\tablesep}}
  @{\hspace{\tablesep}}c@{\hspace{\tablesep}}
}
 & \#class & measure (\%) \\
  \midrule
SR-B   & - & top-10 Acc. \\
SR-T   & - & top-10 Acc. \\
Taxi1500  & 6 & F1 score \\
NER & 7 & F1 score \\
POS & 18 & F1 score \\
  \end{tabular}
  \caption{Information of downstream tasks. \#class: the number of the categories if it is a sequence-level or token-level classification task.}
  \label{evaluation_info_task}
\end{table}

\begin{table}[t]
  \small
	\centering
	\def\tablesep{0.1cm}
\begin{tabular}{
  @{\hspace{\tablesep}}l@{\hspace{\tablesep}}|
  @{\hspace{\tablesep}}r@{\hspace{\tablesep}}
  @{\hspace{\tablesep}}r@{\hspace{\tablesep}}
  @{\hspace{\tablesep}}r@{\hspace{\tablesep}}
  @{\hspace{\tablesep}}r@{\hspace{\tablesep}}
  @{\hspace{\tablesep}}r@{\hspace{\tablesep}}
  @{\hspace{\tablesep}}r@{\hspace{\tablesep}}
  @{\hspace{\tablesep}}r@{\hspace{\tablesep}}
}
 & Latn &Cyrl &Hani &Arab &Deva &other &all \\
 \midrule
SR-B &290 &28 &4 &11 &8 &28 &369 \\
SR-T &64 &10 &3 &5 &2 &14 &98 \\
Taxi1500 &281 &25 &4 &8 &7 &26 &351 \\
NER &104 &17 &4 &10 &5 &24 &164 \\
POS &57 &8 &3 &5 &3 &15 &91 \\
  \end{tabular}
  \caption{The number of languages in each script group in downstream tasks.}
  \label{evaluation_info_script}
\end{table}

The basic information of each downstream task dataset is shown in
Table \ref{evaluation_info_task}. The number of languages in language families and script groups for each downstream task is shown in Table \ref{evaluation_info_family} and \ref{evaluation_info_script} respectively.
We introduce the detailed hyperparameters settings in the following.

For sequence-level retrieval tasks, i.e., \textbf{SR-B} and \textbf{SR-T}, we use English-aligned sentences (up to 500 and 1000 for SR-B and SR-T respectively) from languages that are supported by the model. Different from SR-T, most of the languages in SR-B are low-resource languages. In addition, many languages in SR-B use non-Latin scripts. The retrieval task is performed without any training: we directly use the model to encode all sentences, where each sentence is represented as the average of the contextual embedding at the \textbf{8th} layer. We then compute the top-10 accuracy for each pair.

For sequence-level classification tasks, i.e., \textbf{Taxi1500}, we fine-tune the model with a 6-classes classification head on the English train set and select the best checkpoint using the English development set. We train a model using Adam optimizer for 40 epochs with early stopping. The learning rate is set to 1e-5 and the effective batch size is set to 16 (batch size of 8 and gradient accumulation of 2). The training is done on a single GTX 1080 Ti GPU. We then evaluate the performance in a zero-shot transfer setting by evaluating the fine-tuned model on the test sets of all other languages. The Macro F1 score is reported for each language.

\begin{table*}[h]
    \setlength{\belowcaptionskip}{-0.4cm}
    \scriptsize
    \centering
    \setlength{\tabcolsep}{0.8mm}{}
    \begin{tabular}{l|rrr|rrr|rrr|rrr|rrr}
        \toprule
        & \multicolumn{3}{c}{SR-B} & \multicolumn{3}{c}{SR-T} & \multicolumn{3}{c}{Taxi1500} & \multicolumn{3}{c}{NER} & \multicolumn{3}{c}{POS}\\
        \cmidrule(lr){2-4} \cmidrule(lr){5-7} \cmidrule(lr){8-10} \cmidrule(lr){11-13} \cmidrule(lr){14-16}
        & XLM-R & Glot500 & \modelname & XLM-R & Glot500 & \modelname & XLM-R & Glot500 & \modelname & XLM-R & Glot500 & \modelname & XLM-R & Glot500 & \modelname\\
        \midrule
        indo1319 & 41.9 & \underline{61.6} & \textbf{71.5} & 63.4 & \underline{75.6} & \textbf{77.5} & 48.4 & \underline{61.4} & \textbf{67.7} & 61.0 & \underline{66.0} & \textbf{67.5} & 75.4 & \underline{78.0} & \textbf{78.7} \\
atla1278 & 5.5 & \underline{45.2} & \textbf{56.3} & 29.6 & \underline{50.2} & \textbf{52.6} & 13.3 & \underline{48.2} & \textbf{58.3} & 46.5 & \underline{59.9} & \textbf{60.6} & 24.1 & \underline{60.1} & \textbf{62.6} \\
aust1307 & 14.5 & \underline{47.2} & \textbf{61.6} & 35.3 & \underline{51.0} & \textbf{52.3} & 23.4 & \underline{56.0} & \textbf{62.9} & 49.7 & \textbf{57.6} & \underline{56.8} & 70.1 & \underline{74.6} & \textbf{75.8} \\
turk1311 & 22.3 & \underline{63.3} & \textbf{71.3} & 41.6 & \textbf{70.2} & \underline{65.8} & 30.9 & \underline{62.2} & \textbf{67.0} & 50.7 & \underline{61.9} & \textbf{63.3} & 57.3 & \underline{72.2} & \textbf{73.0} \\
sino1245 & 9.0 & \underline{39.2} & \textbf{46.9} & \underline{62.0} & \textbf{80.5} & 47.7 & 21.9 & \underline{57.4} & \textbf{61.7} & 26.4 & \textbf{37.4} & \underline{35.4} & \underline{22.2} & \textbf{35.5} & 18.2 \\
maya1287 & 3.8 & \underline{20.3} & \textbf{39.6} & - & - & - & 11.1 & \underline{47.8} & \textbf{56.1} & - & - & - & 28.7 & \underline{62.0} & \textbf{64.3} \\
afro1255 & 13.0 & \underline{34.3} & \textbf{47.0} & \underline{41.4} & \textbf{53.0} & 40.0 & 19.3 & \underline{41.4} & \textbf{48.5} & 47.5 & \underline{54.0} & \textbf{58.1} & 54.0 & \textbf{67.2} & \underline{66.3} \\
Other & 14.1 & \underline{36.9} & \textbf{46.2} & 56.7 & \textbf{69.4} & \underline{64.3} & 20.9 & \underline{50.9} & \textbf{56.0} & 50.9 & \underline{56.3} & \textbf{57.5} & 54.1 & \underline{60.8} & \textbf{61.4} \\
All & 19.3 & \underline{47.2} & \textbf{58.1} & 56.6 & \textbf{70.7} & \underline{68.8} & 26.7 & \underline{54.3} & \textbf{61.0} & 55.3 & \underline{61.6} & \textbf{62.8} & 65.6 & \underline{71.8} & \textbf{71.9} \\
        \bottomrule
    \end{tabular}
    \caption{Aggregated performance of \modelname and
    baselines for each major language family. We report the
    average performance for language
    families: \textbf{indo1319}
    (Indo-European), \textbf{atla1278 }
    (Atlantic-Congo), \textbf{aust1307}
    (Austronesian), \textbf{turk1311}
    (Turkic), \textbf{sino1245}
    (Sino-Tibetan), \textbf{maya1287} (Mayan),
    and \textbf{afro1255} (Afro-Asiatic). We collect the
    remaining languages in the group ``\textbf{Other}''. In addition, we also report the average over all languages (group ``\textbf{All}''). 
    \textbf{Bold} (\underline{underlined}): best (second-best) result for each task in each group. \modelname generally performs better than other baselines except on Sino-Tibetan and SR-T.}
    \label{tab:main_results_family}
\end{table*}

For token-level classification tasks, i.e., \textbf{NER} and \textbf{POS}, we fine-tune the model with a suitable classification head (7 for NER and 18 for POS) on the English train set and select the best checkpoint using the English development set. We train each model using Adam optimizer for a maximum of 10 epochs with early stopping. The learning rate is set to 2e-5 and the effective batch size is set to 32 (batch size of 8 and gradient accumulation of 4). The training is done on a single GTX 1080 Ti GPU. We then evaluate the performance in a zero-shot transfer setting by evaluating the fine-tuned model on the test sets of all other languages. The Macro F1 score is reported for each language.

\subsection{Fine-tuning on Indic Group} \label{appendix:indic_fine_tuning}
We fine-tune Glot500-m \citep{imanigooghari-etal-2023-glot500} on a part of indic group languages. Specifically, we randomly sample 10\% of sentences from Glot500-c \citep{imanigooghari-etal-2023-glot500} for 12 indic languages (shown in table \ref{table-indic-group}). Then we create paired data by transliterating each sentence into Latin using \texttt{Uroman} \citep{hermjakob-etal-2018-box}. The other settings are the same in Appendix \ref{Fine-tuning_on_Glot500-c}.

\subsection{Evaluation on Indic Group} \label{appendix:indic_appendix}
The information of three downstream tasks WSTP, NER and CSQA is represented in Table \ref{indic_evaluation_info}. The three tasks are following the same setting of \citet{moosa-etal-2023-transliteration}. For the sentence classification task WSTP, we fine-tune the model with 4-class head on 11 indic languages all at once. We train the model using Adam optimizer for 20 epochs. The learning rate is set to 2e-5 and the effective batch size is set to 256 (batch size 0f 64 and gradient accumulation of 4). The training is done on four GTX 1080 Ti GPUs. We then evaluate the performance in a crosslingual transfer setting by evaluating the fine-tuned model on the test set of 11 indic languages. The accuracy is reported for each language.

For token level task NER, we fine-tune the model with a 7 classification head on 11 indic languages all at once with 20 epochs. The learning rate is set to 2e-5 and the effective batch size is set to 32 (batch size 0f 8 and gradient accumulation of 4). The training is done on a single GTX 1080 Ti GPU. We then evaluate the performance in a crosslingual transfer setting by evaluating the fine-tuned model on the test set of 11 indic languages. The Macro F1 score is reported for each language.

\begin{table}[t]
  \small
	\centering
	\def\tablesep{0.2cm}
\begin{tabular}{
  @{\hspace{\tablesep}}l@{\hspace{\tablesep}}|
  @{\hspace{\tablesep}}r@{\hspace{\tablesep}}
  @{\hspace{\tablesep}}r@{\hspace{\tablesep}}
  @{\hspace{\tablesep}}r@{\hspace{\tablesep}}
  @{\hspace{\tablesep}}c@{\hspace{\tablesep}}
}
 & |lan| & |rows| & \#class & measure (\%) \\
  \midrule
WSTP    & 11 & 403k & 4 & Accuracy \\
NER & 11 & 119k & 7 & F1 score \\
CSQA & 9 & 135k & 4 & Accuracy \\
  \end{tabular}
  \caption{Information of downstream tasks on Indic Group languages. |lan|: languages we evaluate from IndicGlue; \#class: the number of the categories if it is a sequence-level or token-level classification task.}
  \label{indic_evaluation_info}
\end{table}

\section{Futher Fine-grained Analysis}
To further analyze how \frameworkname can influence the crosslinguality of the multilingual model, we additionally report the aggregated results for each language family in Table \ref{tab:main_results_family} and the number of languages that benefit from \frameworkname in Table \ref{tab:whosbetter}. We see similar improvement as we observe for different script groups.  
 
\begin{table}[h]
  \centering
  \small
  \setlength{\tabcolsep}{1.5mm}{}
  \begin{tabular}{l|r|rr}
\toprule
 & |L| & Glot500-m is better & \modelname is better \\
\midrule
SR-B & 369 & 33 & \textbf{336} \\
SR-T & 98 & 43 & \textbf{55} \\
Taxi1500 & 351 & 46 & \textbf{305} \\
NER & 164 & 55 & \textbf{109} \\
POS & 91 & 40 & \textbf{51} \\
\bottomrule
  \end{tabular}
  \caption{Number of languages in each task that benefits from the proposed \frameworkname framework. |L| is the total number of languages for each task.
  \label{tab:whosbetter}}
\end{table}

\section{Complete Results}\seclabel{complete}

We report the complete results for all tasks and language-scripts in Table \ref{bible1}, \ref{bible2} (\textbf{SR-B}), Table \ref{tatoeba} (\textbf{SR-T}), Table \ref{taxi15001}, \ref{taxi15002} (\textbf{Taxi1500}), Table \ref{ner} (\textbf{NER}), and Table \ref{pos} (\textbf{POS}).

\section{Evaluation on Transliterated Data}\seclabel{trans_eval}
We evaluate both Glot500m and \modelname under the common script scenario. Specifically, we transliterate all the data (including data of language written in Latin script, which is consistent with how we fine-tune the model with \frameworkname) from downstream tasks to Latin script. Per-task performance for each script group is shown in Table \ref{tab:uni_bible} (\textbf{SR-B}), Table \ref{tab:uni_tatoeba} (\textbf{SR-T}), Table \ref{tab:uni_taxi1500} (\textbf{Taxi1500}), Table \ref{tab:uni_ner} (\textbf{NER}) and Table \ref{tab:uni_pos} (\textbf{POS}).

The results indicate that the models consistently perform better when the languages are in their original script instead of in Latin (the common script). We hypothesize the major reason is that the models are not being manipulated with vocabulary extension for transliterated data. Nevertheless, we observe from the results that \frameworkname-uni generally outperforms Glot500m-uni across all tasks. This indicates our \frameworkname framework is also effective for common script scenarios. 

\begin{table}[!h]
    \setlength{\belowcaptionskip}{-0.4cm}
    \scriptsize
    \centering
    \setlength{\tabcolsep}{1.5mm}{}
    \begin{tabular}{l|rrrr}
        \toprule
         & Glot500-m-uni & Glot500-m & \modelname-uni & \modelname \\
        \midrule
Latn & 38.7 & 45.1 & \underline{53.8} & \textbf{57.4} \\
Cyrl & 19.2 & \underline{60.3} & 47.0 & \textbf{69.0} \\
Hani & 5.7 & \textbf{43.4} & 10.4 & \underline{39.8} \\
Arab & 5.9 & \underline{56.4} & 22.9 & \textbf{61.4} \\
Deva & 9.9 & \underline{60.3} & 32.7 & \textbf{66.8} \\
Other & 5.4 & \underline{49.0} & 18.0 & \textbf{53.6} \\
All & 32.7 & 47.2 & \underline{48.7} & \textbf{58.1} \\
        \bottomrule
    \end{tabular}
    \caption{Performance Glot500-m and \modelname on the original and transliterated (into Latin) evaluation dataset of \textbf{SR-B}. We use Glot500-m and \modelname (resp. Glot500-m-uni and \modelname-uni) to refer to the model performing evaluation on the original-script (resp. transliterated) dataset. We group the performance by scripts (for Glot500-m-uni and \modelname-uni, the script indicates the original script of the languages).} 
    \label{tab:uni_bible}
\end{table}

\begin{table}[!h]
    \setlength{\belowcaptionskip}{-0.4cm}
    \scriptsize
    \centering
    \setlength{\tabcolsep}{1.5mm}{}
    \begin{tabular}{l|rrrr}
        \toprule
 & Glot500-m-uni & Glot500-m & \modelname-uni & \modelname \\
\midrule
Latn & 58.7 & \underline{69.1} & 68.8 & \textbf{73.0} \\
Cyrl & 28.0 & \textbf{74.4} & 50.4 & \underline{69.7} \\
Hani & 4.5 & \textbf{80.5} & 6.2 & \underline{47.7} \\
Arab & 6.7 & \textbf{71.8} & 16.1 & \underline{56.3} \\
Deva & 16.0 & \textbf{81.8} & 37.4 & \underline{71.9} \\
Other & 7.0 & \textbf{71.1} & 15.7 & \underline{57.6} \\
All & 43.0 & \textbf{70.7} & 54.1 & \underline{68.8} \\
        \bottomrule
    \end{tabular}
    \caption{Performance Glot500-m and \modelname on the original and transliterated (into Latin) evaluation dataset of \textbf{SR-T}. We use Glot500-m and \modelname (resp. Glot500-m-uni and \modelname-uni) to refer to the model performing evaluation on the original-script (resp. transliterated) dataset. We group the performance by scripts (for Glot500-m-uni and \modelname-uni, the script indicates the original script of the languages).} 
    \label{tab:uni_tatoeba}
\end{table}

\begin{table}[!h]
    \setlength{\belowcaptionskip}{-0.4cm}
    \scriptsize
    \centering
    \setlength{\tabcolsep}{1.5mm}{}
    \begin{tabular}{l|rrrr}
        \toprule
 & Glot500-m-uni & Glot500-m & \modelname-uni & \modelname \\
\midrule
Latn & 50.5 & \underline{52.6} & 48.4 & \textbf{59.8} \\
Cyrl & 29.6 & \underline{59.8} & 30.6 & \textbf{63.6} \\
Hani & 6.9 & \underline{68.2} & 5.2 & \textbf{70.1} \\
Arab & 15.4 & \underline{60.8} & 15.6 & \textbf{66.5} \\
Deva & 21.6 & \underline{66.6} & 24.1 & \textbf{73.2} \\
Other & 11.5 & \underline{59.5} & 14.7 & \textbf{65.2} \\
All & 44.2 & \underline{54.3} & 42.9 & \textbf{61.0} \\
        \bottomrule
    \end{tabular}
    \caption{Performance Glot500-m and \modelname on the original and transliterated (into Latin) evaluation dataset of \textbf{Taxi1500}. We use Glot500-m and \modelname (resp. Glot500-m-uni and \modelname-uni) to refer to the model performing evaluation on the original-script (resp. transliterated) dataset. We group the performance by scripts (for Glot500-m-uni and \modelname-uni, the script indicates the original script of the languages).} 
    \label{tab:uni_taxi1500}
\end{table}

\begin{table}[!h]
    \setlength{\belowcaptionskip}{-0.4cm}
    \scriptsize
    \centering
    \setlength{\tabcolsep}{1.5mm}{}
    \begin{tabular}{l|rrrr}
        \toprule
 & Glot500-m-uni & Glot500-m & \modelname-uni & \modelname \\
\midrule
Latn & 64.3 & 66.1 & \underline{66.2} & \textbf{67.3} \\
Cyrl & 49.5 & \underline{65.3} & 57.5 & \textbf{66.2} \\
Hani & 10.6 & \textbf{22.2} & 10.0 & \underline{21.9} \\
Arab & 14.5 & \underline{53.4} & 21.2 & \textbf{57.7} \\
Deva & 14.0 & \underline{56.2} & 29.1 & \textbf{58.9} \\
Other & 16.6 & \textbf{50.4} & 24.6 & \textbf{50.4} \\
All & 49.9 & \underline{61.6} & 54.0 & \textbf{62.8} \\
        \bottomrule
    \end{tabular}
    \caption{Performance Glot500-m and \modelname on the original and transliterated (into Latin) evaluation dataset of \textbf{NER}. We use Glot500-m and \modelname (resp. Glot500-m-uni and \modelname-uni) to refer to the model performing evaluation on the original-script (resp. transliterated) dataset. We group the performance by scripts (for Glot500-m-uni and \modelname-uni, the script indicates the original script of the languages).} 
    \label{tab:uni_ner}
\end{table}

\begin{table}[!h]
    \setlength{\belowcaptionskip}{-0.4cm}
    \scriptsize
    \centering
    \setlength{\tabcolsep}{1.5mm}{}
    \begin{tabular}{l|rrrr}
        \toprule
 & Glot500-m-uni & Glot500-m & \modelname-uni & \modelname \\
\midrule
Latn & 70.0 & \underline{74.4} & 72.5 & \textbf{75.7} \\
Cyrl & 51.8 & \underline{79.3} & 63.1 & \textbf{79.5} \\
Hani & 22.7 & \textbf{35.5} & \underline{23.4} & 18.2 \\
Arab & 28.1 & \underline{68.8} & 46.0 & \textbf{69.3} \\
Deva & 33.7 & \underline{59.8} & 46.4 & \textbf{60.8} \\
Other & 32.5 & \textbf{68.8} & 41.2 & \underline{67.1} \\
All & 57.1 & \underline{71.8} & 62.6 & \textbf{71.9} \\
        \bottomrule
    \end{tabular}
    \caption{Performance Glot500-m and \modelname on the original and transliterated (into Latin) evaluation dataset of \textbf{POS}. We use Glot500-m and \modelname (resp. Glot500-m-uni and \modelname-uni) to refer to the model performing evaluation on the original-script (resp. transliterated) dataset. We group the performance by scripts (for Glot500-m-uni and \modelname-uni, the script indicates the original script of the languages).} 
    \label{tab:uni_pos}
\end{table}

\section{Representation Visualization}\label{appendix:visual_appendix}
\begin{figure*}[h!]
\centering
\subfigure[layer 1]{
\begin{minipage}[t]{0.3\linewidth}
\centering
\includegraphics[width=1.6in]{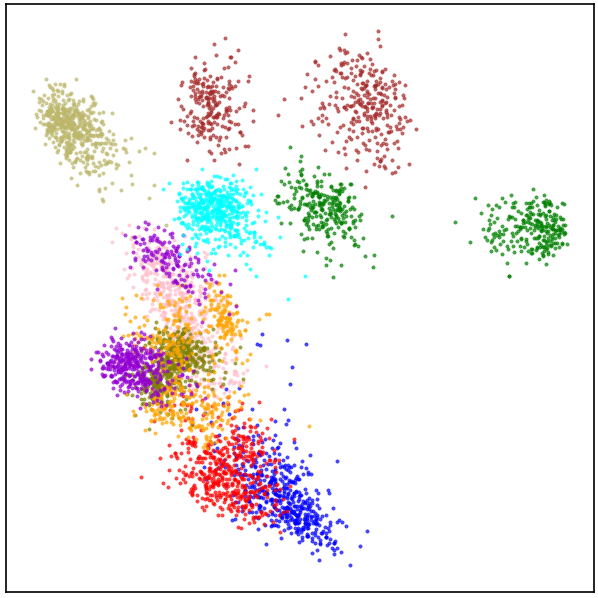}
\end{minipage}%
}%
\subfigure[layer 2]{
\begin{minipage}[t]{0.3\linewidth}
\centering
\includegraphics[width=1.6in]{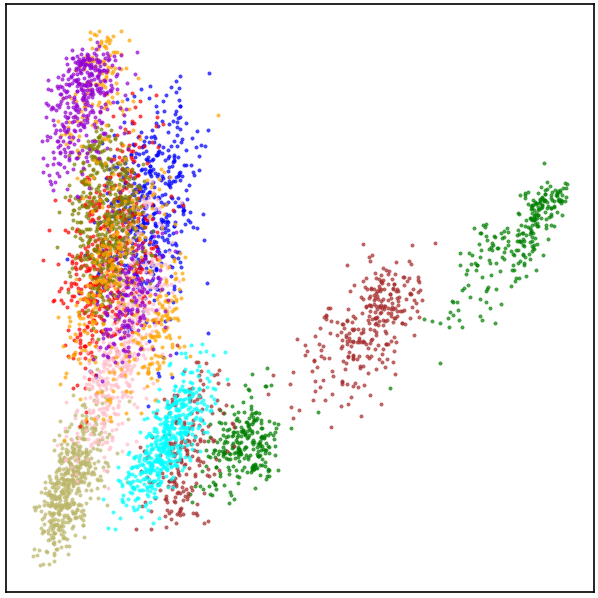}
\end{minipage}%
}%
\subfigure[layer 3]{
\begin{minipage}[t]{0.3\linewidth}
\centering
\includegraphics[width=1.6in]{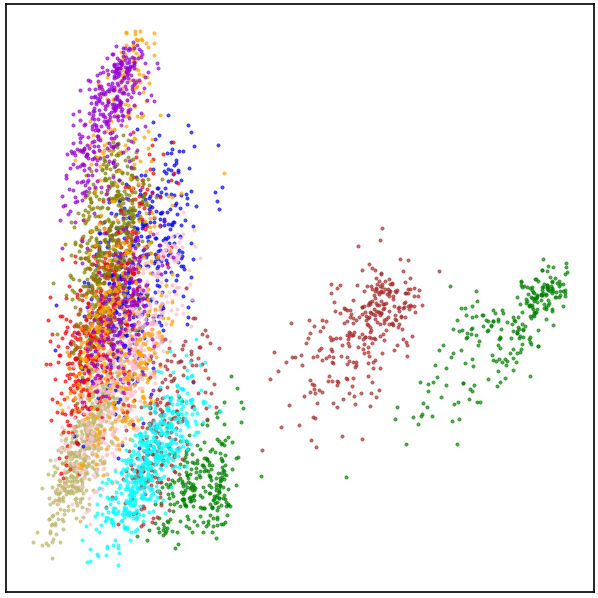}
\end{minipage}%
}%

\subfigure[layer 4]{
\begin{minipage}[t]{0.3\linewidth}
\centering
\includegraphics[width=1.6in]{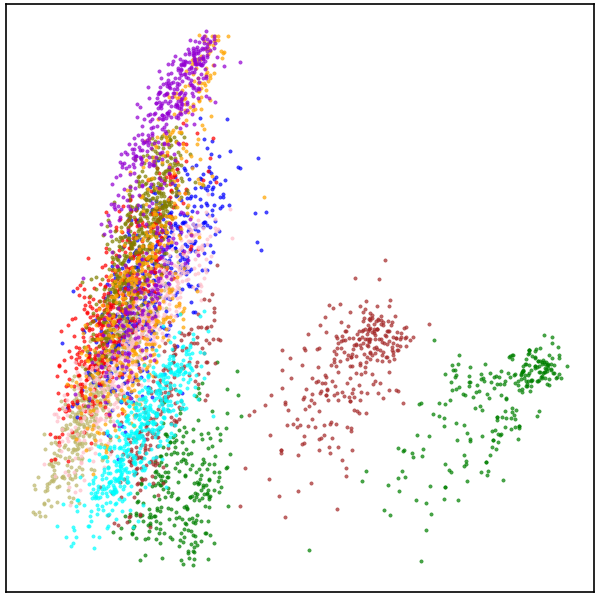}
\end{minipage}%
}%
\subfigure[layer 5]{
\begin{minipage}[t]{0.3\linewidth}
\centering
\includegraphics[width=1.6in]{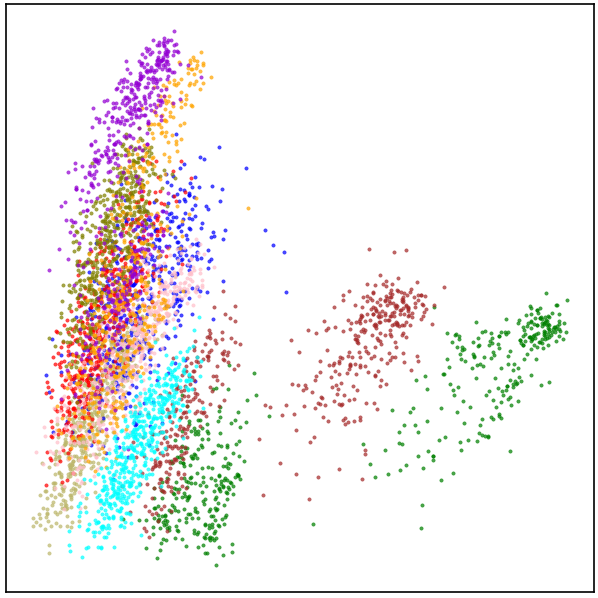}
\end{minipage}%
}%
\subfigure[layer 6]{
\begin{minipage}[t]{0.3\linewidth}
\centering
\includegraphics[width=1.6in]{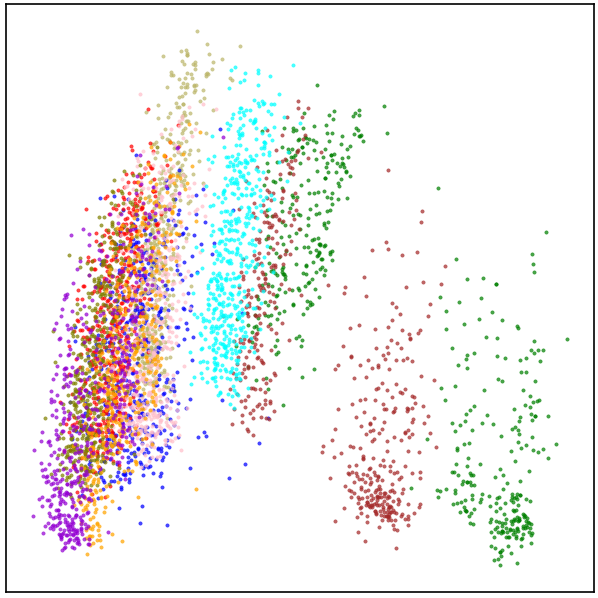}
\end{minipage}%
}%

\subfigure[layer 7]{
\begin{minipage}[t]{0.3\linewidth}
\centering
\includegraphics[width=1.6in]{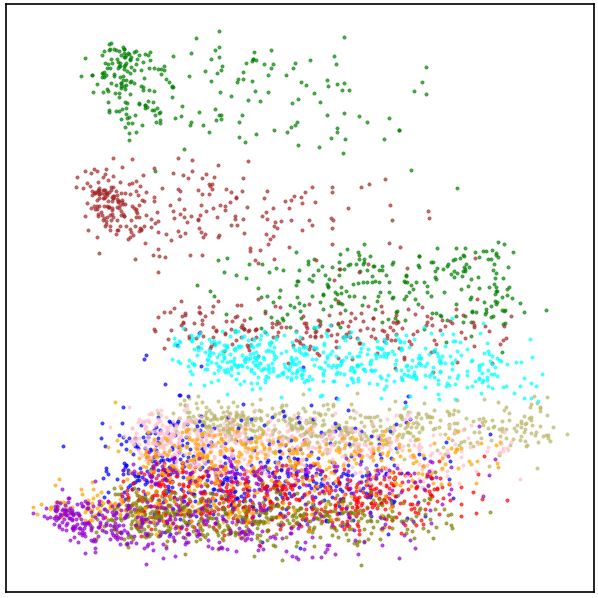}
\end{minipage}%
}%
\subfigure[layer 8]{
\begin{minipage}[t]{0.3\linewidth}
\centering
\includegraphics[width=1.6in]{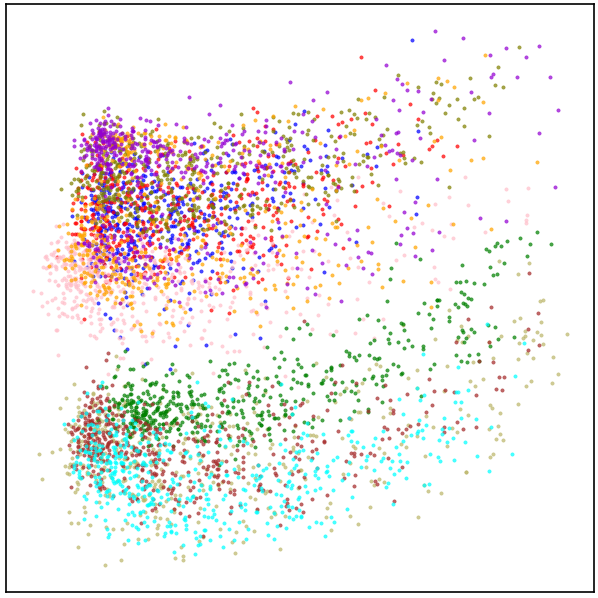}
\end{minipage}%
}%
\subfigure[layer 9]{
\begin{minipage}[t]{0.3\linewidth}
\centering
\includegraphics[width=1.6in]{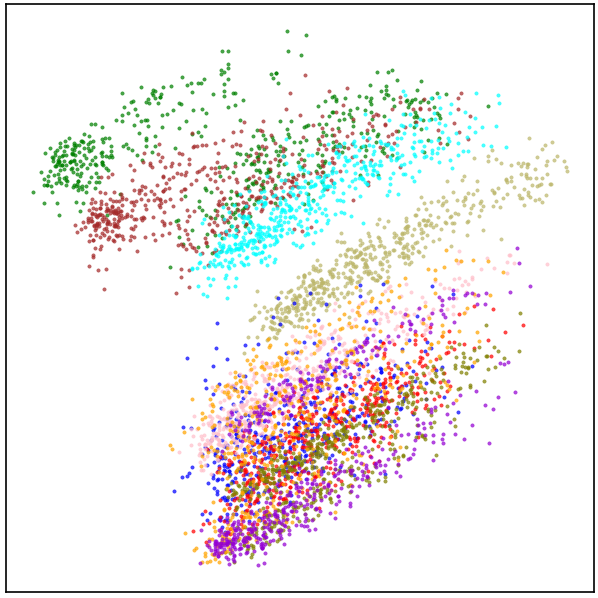}
\end{minipage}%
}%

\subfigure[layer 10]{
\begin{minipage}[t]{0.3\linewidth}
\centering
\includegraphics[width=1.6in]{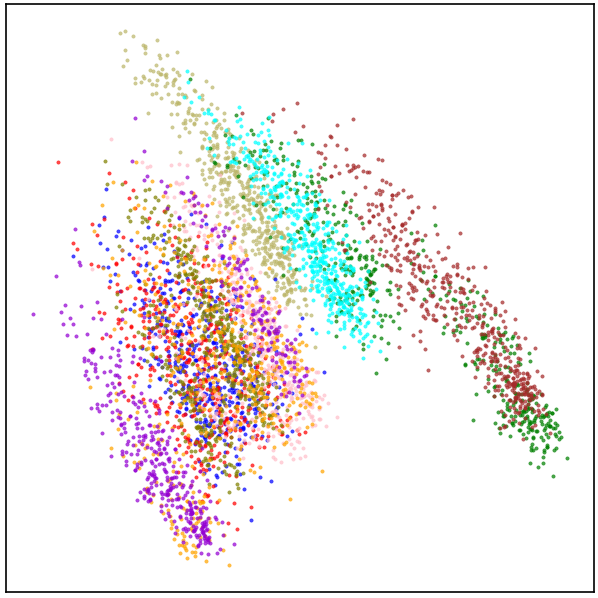}
\end{minipage}%
}%
\subfigure[layer 11]{
\begin{minipage}[t]{0.3\linewidth}
\centering
\includegraphics[width=1.6in]{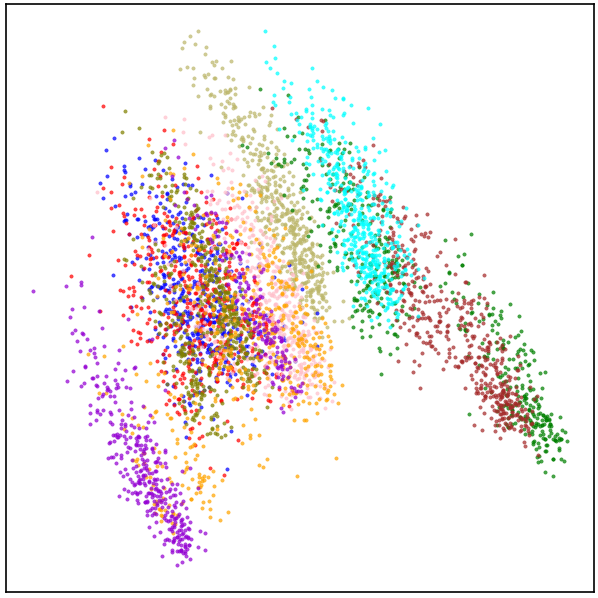}
\end{minipage}%
}%
\subfigure[layer 12]{
\begin{minipage}[t]{0.3\linewidth}
\centering
\includegraphics[width=1.6in]{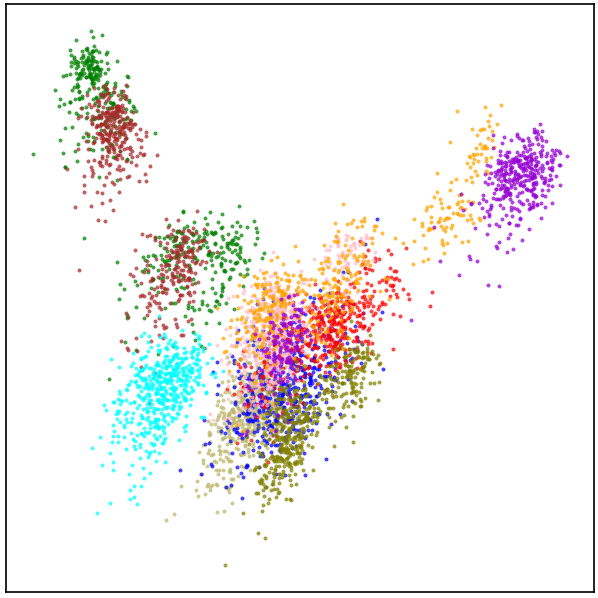}
\end{minipage}%
}%
\centering
\caption{Visualizations of sentence representations from 
all layers (mean-pooling the contextualized token embeddings)
of \modelname. The original dimension is 768 and we use PCA
to select the first two principal components. Each point
corresponds to a sentence. Different colors indicate distinct scripts.}
\label{appfig:Furina_conmbied_figure}
\end{figure*}

\begin{figure*}[h!]
\centering
\subfigure[layer 1]{
\begin{minipage}[t]{0.3\linewidth}
\centering
\includegraphics[width=1.6in]{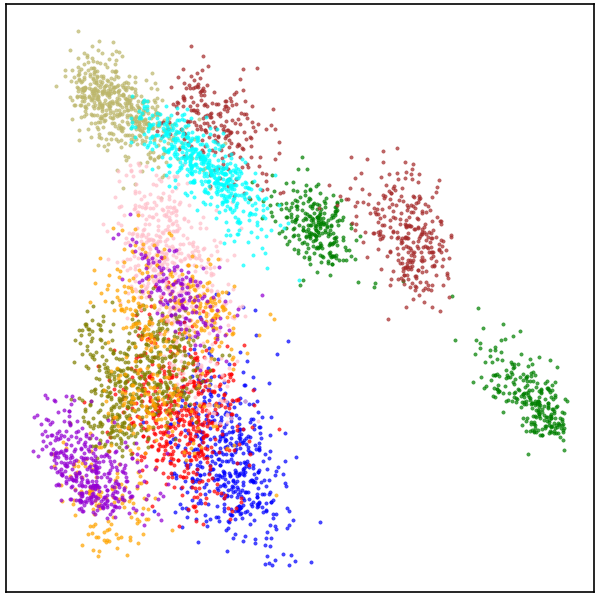}
\end{minipage}%
}%
\subfigure[layer 2]{
\begin{minipage}[t]{0.3\linewidth}
\centering
\includegraphics[width=1.6in]{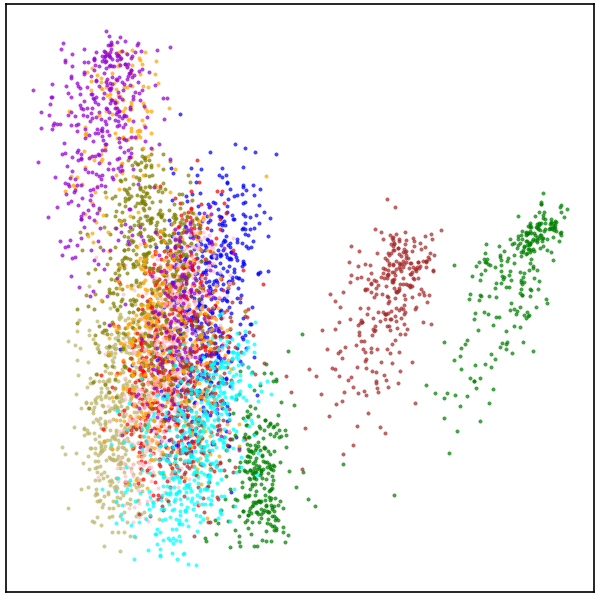}
\end{minipage}%
}%
\subfigure[layer 3]{
\begin{minipage}[t]{0.3\linewidth}
\centering
\includegraphics[width=1.6in]{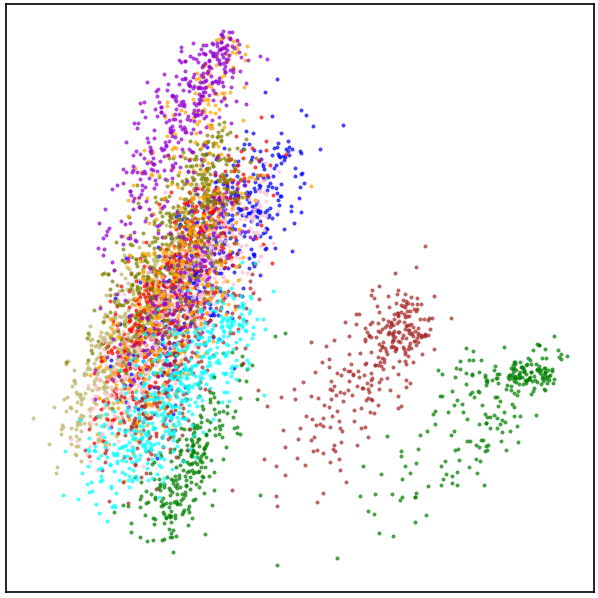}
\end{minipage}%
}%

\subfigure[layer 4]{
\begin{minipage}[t]{0.3\linewidth}
\centering
\includegraphics[width=1.6in]{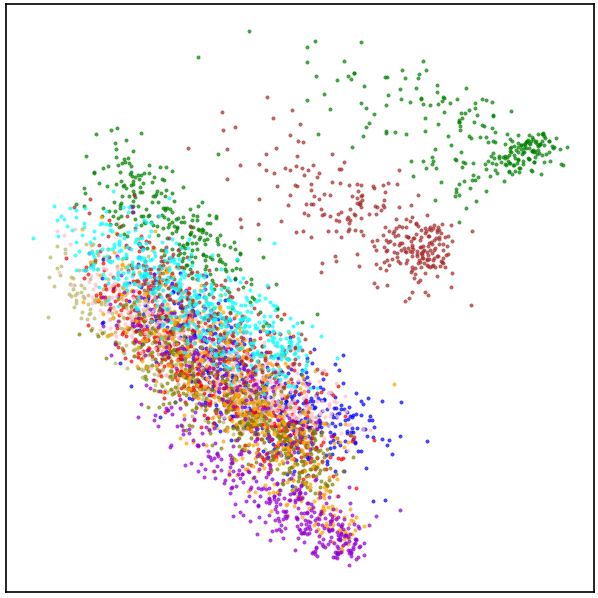}
\end{minipage}%
}%
\subfigure[layer 5]{
\begin{minipage}[t]{0.3\linewidth}
\centering
\includegraphics[width=1.6in]{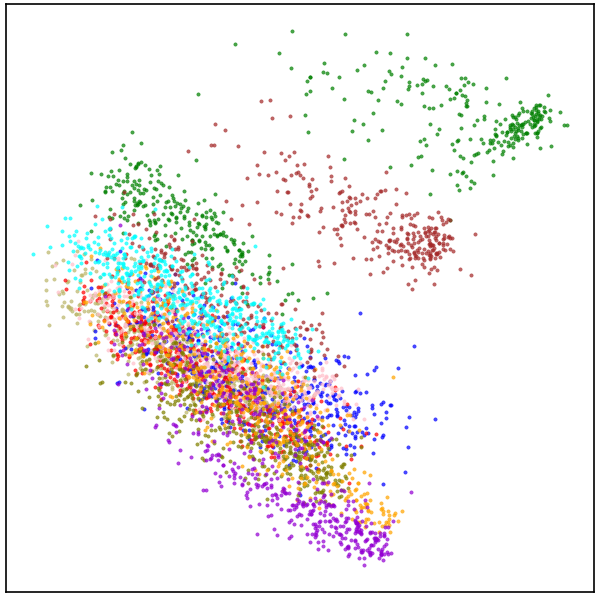}
\end{minipage}%
}%
\subfigure[layer 6]{
\begin{minipage}[t]{0.3\linewidth}
\centering
\includegraphics[width=1.6in]{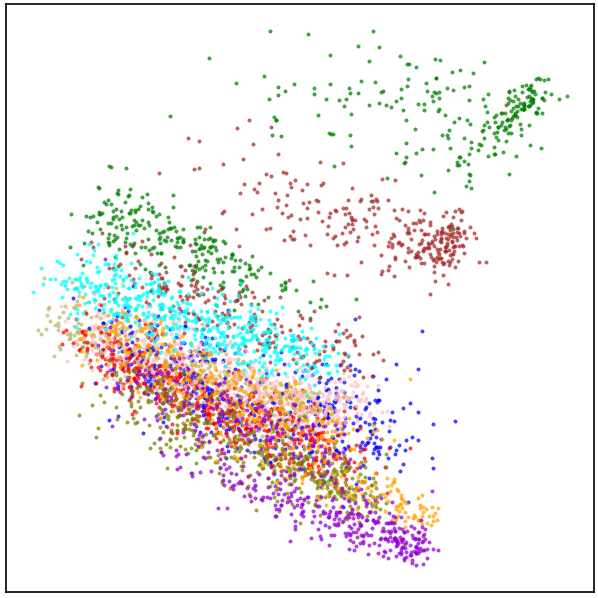}
\end{minipage}%
}%

\subfigure[layer 7]{
\begin{minipage}[t]{0.3\linewidth}
\centering
\includegraphics[width=1.6in]{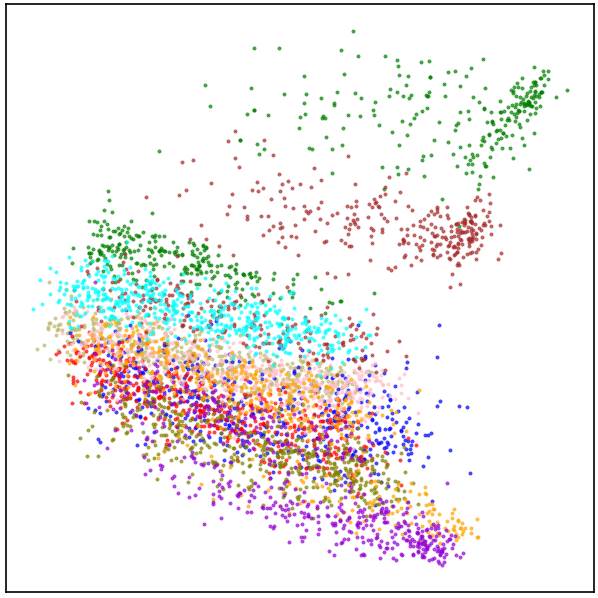}
\end{minipage}%
}%
\subfigure[layer 8]{
\begin{minipage}[t]{0.3\linewidth}
\centering
\includegraphics[width=1.6in]{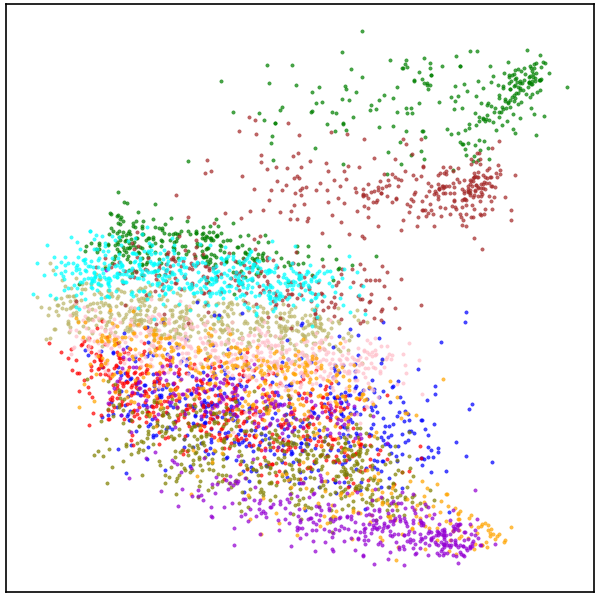}
\end{minipage}%
}%
\subfigure[layer 9]{
\begin{minipage}[t]{0.3\linewidth}
\centering
\includegraphics[width=1.6in]{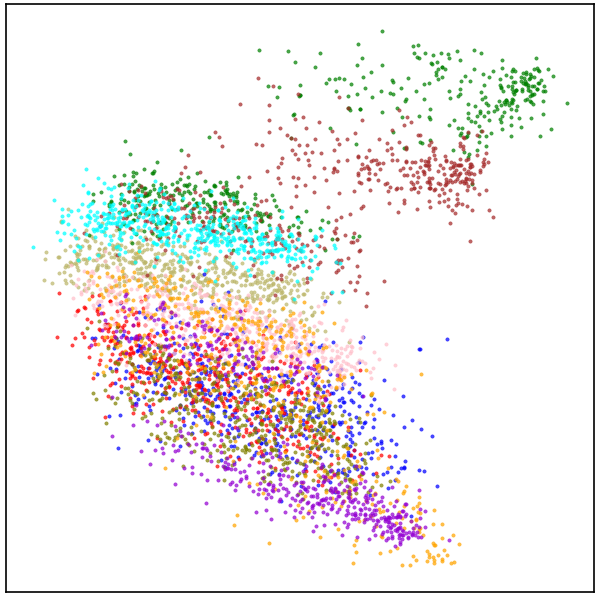}
\end{minipage}%
}%

\subfigure[layer 10]{
\begin{minipage}[t]{0.3\linewidth}
\centering
\includegraphics[width=1.6in]{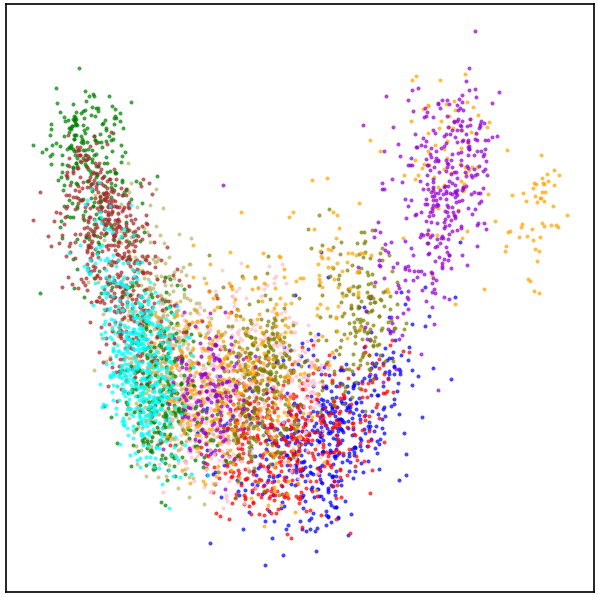}
\end{minipage}%
}%
\subfigure[layer 11]{
\begin{minipage}[t]{0.3\linewidth}
\centering
\includegraphics[width=1.6in]{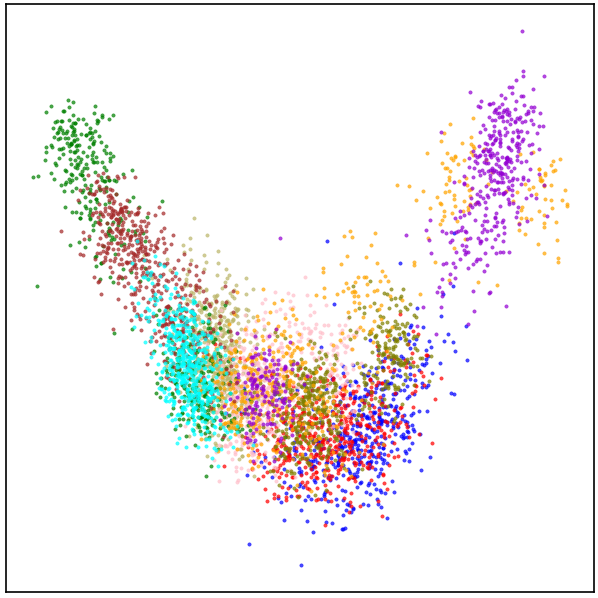}
\end{minipage}%
}%
\subfigure[layer 12]{
\begin{minipage}[t]{0.3\linewidth}
\centering
\includegraphics[width=1.6in]{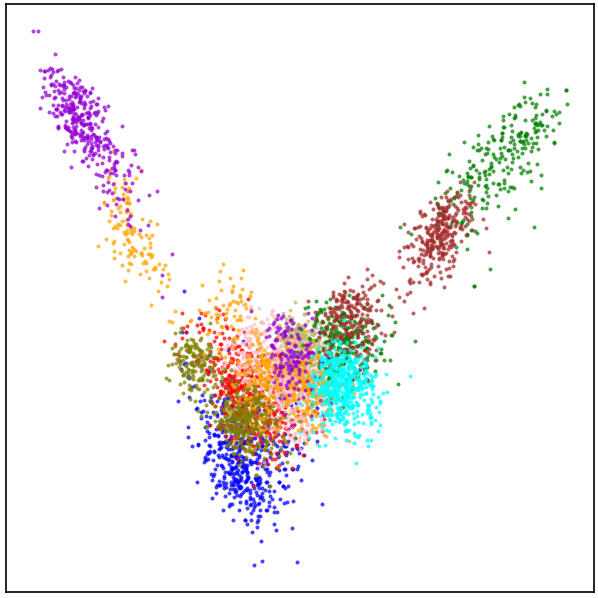}
\end{minipage}%
}%
\centering
\caption{Visualizations of sentence representations from all layers (mean-pooling the contextualized token embeddings) of Glot500-m. The original dimension is 768 and we use PCA to select the first two principal components. Points indicate sentence representations and different colors indicate distinct scripts.}
\label{appfig:Glot500_conmbied_figure}
\end{figure*}

We visualized sentence representations from all layers of two models using PCA. Figure \ref{appfig:Glot500_conmbied_figure} and Figure \ref{appfig:Furina_conmbied_figure} present the sentence representations of Glot500-m and \modelname respectively.

\begin{table*}
\centering
\small
\resizebox{\textwidth}{!}{
\begin{tabular}{lrrr|lrrr|lrrr}
    \toprule
    Language-script & XLM-R & Glot500-m & \modelname & Language-script & XLM-R & Glot500-m & \modelname & Language-script & XLM-R & Glot500-m & \modelname \\
    \midrule
ace\_Latn & 4.4 & \underline{53.4} & \textbf{72.0} & ach\_Latn & 4.4 & \underline{40.0} & \textbf{50.0} & acr\_Latn & 2.6 & \underline{25.4} & \textbf{48.2} \\
afr\_Latn & \underline{76.8} & 69.4 & \textbf{81.4} & agw\_Latn & 5.8 & \underline{36.0} & \textbf{52.8} & ahk\_Latn & 3.0 & \underline{3.2} & \textbf{8.0} \\
aka\_Latn & 5.0 & \textbf{57.0} & \underline{53.4} & aln\_Latn & \underline{67.8} & 67.6 & \textbf{73.0} & als\_Latn & 51.4 & \underline{55.8} & \textbf{56.4} \\
alt\_Cyrl & 12.6 & \underline{50.8} & \textbf{59.2} & alz\_Latn & 4.6 & \underline{34.6} & \textbf{44.6} & amh\_Ethi & 35.4 & \textbf{52.8} & \underline{43.8} \\
aoj\_Latn & 5.0 & \underline{20.4} & \textbf{32.8} & arb\_Arab & 7.0 & \underline{14.6} & \textbf{35.4} & arn\_Latn & 4.8 & \underline{28.4} & \textbf{36.4} \\
ary\_Arab & 2.8 & \underline{15.2} & \textbf{34.2} & arz\_Arab & 5.4 & \underline{24.8} & \textbf{45.4} & asm\_Beng & 26.2 & \underline{66.6} & \textbf{73.4} \\
ayr\_Latn & 4.8 & \underline{52.8} & \textbf{63.8} & azb\_Arab & 7.4 & \underline{72.4} & \textbf{74.2} & aze\_Latn & 71.0 & \underline{73.0} & \textbf{76.0} \\
\bottomrule
    \end{tabular}
}
    \caption{Top-10 accuracy of baselines and \modelname on \textbf{SR-B} (Part I).}\label{bible1}
\end{table*}

\begin{table*}
\centering
\small
\resizebox{\textwidth}{!}{
    \begin{tabular}{lrrr|lrrr|lrrr}
    \toprule
    Language-script & XLM-R & Glot500-m & \modelname & Language-script & XLM-R & Glot500-m & \modelname & Language-script & XLM-R & Glot500-m & \modelname \\
    \midrule
nyn\_Latn & 4.4 & \underline{51.8} & \textbf{60.4} & nyy\_Latn & 3.0 & \underline{25.6} & \textbf{40.6} & nzi\_Latn & 3.2 & \underline{47.2} & \textbf{56.4} \\
ori\_Orya & 42.6 & \underline{56.0} & \textbf{75.2} & ory\_Orya & 31.4 & \underline{53.4} & \textbf{63.8} & oss\_Cyrl & 4.2 & \underline{54.8} & \textbf{63.8} \\
ote\_Latn & 3.6 & \underline{18.0} & \textbf{40.0} & pag\_Latn & 8.0 & \underline{61.2} & \textbf{76.8} & pam\_Latn & 8.2 & \underline{49.8} & \textbf{77.0} \\
pan\_Guru & 43.2 & \underline{48.8} & \textbf{58.0} & pap\_Latn & 12.4 & \underline{72.4} & \textbf{79.8} & pau\_Latn & 4.4 & \underline{29.8} & \textbf{46.4} \\
pcm\_Latn & 13.6 & \underline{66.8} & \textbf{73.2} & pdt\_Latn & 9.2 & \underline{68.6} & \textbf{77.0} & pes\_Arab & \underline{69.4} & \textbf{80.6} & 69.0 \\
pis\_Latn & 6.4 & \underline{57.2} & \textbf{74.6} & pls\_Latn & 5.0 & \underline{34.4} & \textbf{56.2} & plt\_Latn & 26.6 & \underline{59.8} & \textbf{66.8} \\
poh\_Latn & 3.4 & \underline{15.2} & \textbf{33.4} & pol\_Latn & \textbf{79.2} & 63.8 & \underline{78.6} & pon\_Latn & 5.6 & \underline{21.6} & \textbf{36.2} \\
\bottomrule
    \end{tabular}
}
    \caption{Top-10 accuracy of baselines and \modelname on \textbf{SR-B} (Part II).}\label{bible2}
\end{table*}
\begin{table*}
\centering
\small
\resizebox{\textwidth}{!}{
    \begin{tabular}{lrrr|lrrr|lrrr}
    \toprule
    Language-script & XLM-R & Glot500-m & \modelname & Language-script & XLM-R & Glot500-m & \modelname & Language-script & XLM-R & Glot500-m & \modelname \\
    \midrule
afr\_Latn & 71.9 & \underline{81.1} & \textbf{84.7} & amh\_Ethi & 35.1 & \textbf{44.6} & \underline{44.0} & ara\_Arab & \underline{59.2} & \textbf{64.2} & 46.6 \\
arz\_Arab & 32.5 & \textbf{63.5} & \underline{40.3} & ast\_Latn & 59.8 & \underline{87.4} & \textbf{88.2} & aze\_Latn & 62.6 & \underline{79.9} & \textbf{81.9} \\
bel\_Cyrl & 70.0 & \textbf{81.4} & \underline{79.0} & ben\_Beng & 54.1 & \underline{69.4} & \textbf{71.3} & bos\_Latn & 78.5 & \underline{92.4} & \textbf{94.1} \\
bre\_Latn & 10.3 & \underline{19.9} & \textbf{21.7} & bul\_Cyrl & 84.4 & \textbf{86.7} & \textbf{86.7} & cat\_Latn & 72.8 & \underline{78.7} & \textbf{84.6} \\
cbk\_Latn & 33.2 & \underline{49.4} & \textbf{59.8} & ceb\_Latn & 15.2 & \underline{41.3} & \textbf{42.5} & ces\_Latn & 71.1 & \underline{75.1} & \textbf{78.2} \\
cmn\_Hani & \underline{79.5} & \textbf{85.6} & 39.7 & csb\_Latn & 21.3 & \underline{40.3} & \textbf{68.4} & cym\_Latn & 45.7 & \textbf{55.7} & \underline{53.9} \\
dan\_Latn & \underline{91.9} & 91.5 & \textbf{94.7} & deu\_Latn & \underline{95.9} & 95.0 & \textbf{96.5} & dtp\_Latn & 5.6 & \textbf{21.1} & \underline{19.8} \\
\bottomrule
    \end{tabular}
}
    \caption{Top-10 accuracy of baselines and \modelname on \textbf{SR-T}.}\label{tatoeba}
\end{table*}
\begin{table*}
\centering
\small
\resizebox{\textwidth}{!}{
    \begin{tabular}{lrrr|lrrr|lrrr}
    \toprule
    Language-script & XLM-R & Glot500-m & \modelname & Language-script & XLM-R & Glot500-m & \modelname & Language-script & XLM-R & Glot500-m & \modelname \\
    \midrule
ace\_Latn & 13.4 & \underline{65.6} & \textbf{70.5} & ach\_Latn & 10.9 & \underline{39.6} & \textbf{55.9} & acr\_Latn & 8.8 & \underline{51.7} & \textbf{67.7} \\
afr\_Latn & \underline{65.7} & 65.5 & \textbf{67.0} & agw\_Latn & 13.9 & \textbf{63.0} & \underline{60.6} & ahk\_Latn & \textbf{9.3} & 8.9 & \underline{9.1} \\
aka\_Latn & 9.1 & \underline{47.8} & \textbf{55.7} & aln\_Latn & 53.8 & \underline{59.3} & \textbf{62.3} & als\_Latn & 57.8 & \textbf{60.9} & \textbf{60.9} \\
alt\_Cyrl & 25.4 & \underline{46.9} & \textbf{50.4} & alz\_Latn & 11.8 & \underline{32.4} & \textbf{49.3} & amh\_Ethi & 9.3 & \textbf{14.1} & \underline{10.1} \\
aoj\_Latn & 12.2 & \underline{45.2} & \textbf{56.4} & arn\_Latn & 9.1 & \underline{48.5} & \textbf{50.8} & ary\_Arab & 14.5 & \underline{39.4} & \textbf{43.4} \\
arz\_Arab & 21.9 & \underline{39.3} & \textbf{45.6} & asm\_Beng & 47.3 & \underline{60.9} & \textbf{69.6} & ayr\_Latn & 7.7 & \underline{60.7} & \textbf{72.3} \\
azb\_Arab & 16.1 & \underline{60.5} & \textbf{67.1} & aze\_Latn & 64.6 & \underline{68.1} & \textbf{73.5} & bak\_Cyrl & 22.6 & \underline{64.0} & \textbf{71.6} \\
\bottomrule
    \end{tabular}
}
    \caption{F1 scores of baselines and \modelname on \textbf{Taxi1500} (Part I).}\label{taxi15001}
\end{table*}

\begin{table*}
\centering
\small
\resizebox{\textwidth}{!}{
    \begin{tabular}{lrrr|lrrr|lrrr}
    \toprule
    Language-script & XLM-R & Glot500-m & \modelname & Language-script & XLM-R & Glot500-m & \modelname & Language-script & XLM-R & Glot500-m & \modelname \\
    \midrule
pap\_Latn & 36.3 & \underline{60.9} & \textbf{76.3} & pau\_Latn & 15.6 & \underline{42.2} & \textbf{49.7} & pcm\_Latn & 31.8 & \textbf{70.6} & \underline{60.9} \\
pdt\_Latn & 18.1 & \underline{57.6} & \textbf{66.3} & pes\_Arab & 72.6 & \textbf{74.3} & \underline{73.3} & pis\_Latn & 12.5 & \underline{68.1} & \textbf{69.1} \\
pls\_Latn & 16.2 & \underline{54.9} & \textbf{60.6} & plt\_Latn & 32.3 & \underline{49.8} & \textbf{59.4} & poh\_Latn & 12.7 & \textbf{55.9} & \underline{55.7} \\
pol\_Latn & \underline{68.8} & 60.9 & \textbf{76.7} & pon\_Latn & 7.9 & \underline{62.7} & \textbf{64.8} & por\_Latn & \underline{73.4} & 62.6 & \textbf{74.7} \\
prk\_Latn & 11.2 & \underline{63.3} & \textbf{67.2} & prs\_Arab & \underline{74.4} & 69.8 & \textbf{78.5} & pxm\_Latn & 11.5 & \underline{44.4} & \textbf{48.7} \\
qub\_Latn & 10.1 & \underline{64.3} & \textbf{70.6} & quc\_Latn & 15.3 & \underline{54.3} & \textbf{61.7} & qug\_Latn & 12.0 & \underline{67.1} & \textbf{73.6} \\
quh\_Latn & 12.1 & \underline{70.1} & \textbf{75.2} & quw\_Latn & 11.2 & \underline{58.9} & \textbf{65.9} & quy\_Latn & 11.1 & \underline{71.4} & \textbf{74.4} \\
\bottomrule
    \end{tabular}
}
    \caption{F1 scores of baselines and \modelname on \textbf{Taxi1500} (Part II).}\label{taxi15002}
\end{table*}
\begin{table*}
\centering
\small
\resizebox{\textwidth}{!}{
    \begin{tabular}{lrrr|lrrr|lrrr}
    \toprule
    Language-script & XLM-R & Glot500-m & \modelname & Language-script & XLM-R & Glot500-m & \modelname & Language-script & XLM-R & Glot500-m & \modelname \\
    \midrule
ace\_Latn & 33.7 & \textbf{43.2} & \textbf{43.2} & afr\_Latn & \underline{75.7} & 74.9 & \textbf{76.1} & als\_Latn & 61.8 & \textbf{80.2} & \underline{76.0} \\
amh\_Ethi & \textbf{41.8} & \underline{41.7} & 40.0 & ara\_Arab & 45.4 & \underline{55.9} & \textbf{66.0} & arg\_Latn & 73.7 & \underline{75.8} & \textbf{79.9} \\
arz\_Arab & 48.0 & \underline{51.4} & \textbf{59.0} & asm\_Beng & 53.3 & \textbf{66.7} & \underline{65.5} & ast\_Latn & 80.2 & \underline{83.1} & \textbf{85.3} \\
aym\_Latn & 36.0 & \underline{44.7} & \textbf{46.1} & aze\_Latn & \underline{63.6} & \underline{63.6} & \textbf{67.7} & bak\_Cyrl & 36.6 & \textbf{59.0} & \underline{58.6} \\
bar\_Latn & 57.5 & \underline{72.4} & \textbf{75.7} & bel\_Cyrl & 73.2 & \underline{73.9} & \textbf{75.1} & ben\_Beng & 65.5 & \underline{73.9} & \textbf{74.9} \\
bih\_Deva & 50.0 & \textbf{58.9} & \underline{57.1} & bod\_Tibt & 0.0 & \textbf{34.6} & \underline{33.5} & bos\_Latn & \textbf{74.5} & 72.0 & \underline{74.2} \\
bre\_Latn & 59.5 & \underline{62.7} & \textbf{65.1} & bul\_Cyrl & \underline{77.2} & \textbf{77.7} & 77.1 & cat\_Latn & 81.8 & \underline{83.7} & \textbf{84.0} \\
\bottomrule
    \end{tabular}
}
    \caption{F1 scores of baselines and \modelname on \textbf{NER}.}\label{ner}
\end{table*}
\begin{table*}
\centering
\small
\resizebox{\textwidth}{!}{
    \begin{tabular}{lrrr|lrrr|lrrr}
    \toprule
    Language-script & XLM-R & Glot500-m & \modelname & Language-script & XLM-R & Glot500-m & \modelname & Language-script & XLM-R & Glot500-m & \modelname \\
    \midrule
afr\_Latn & \textbf{89.3} & \underline{87.7} & 86.7 & ajp\_Arab & 63.0 & \textbf{73.1} & \underline{70.1} & aln\_Latn & \underline{54.1} & 49.7 & \textbf{54.2} \\
amh\_Ethi & \underline{63.9} & \textbf{65.9} & 63.6 & ara\_Arab & \textbf{67.9} & \underline{65.7} & 65.3 & bam\_Latn & 25.1 & \underline{39.5} & \textbf{49.7} \\
bel\_Cyrl & \textbf{86.0} & \underline{85.7} & 85.2 & ben\_Beng & 82.0 & \textbf{83.9} & \underline{82.2} & bre\_Latn & \underline{61.3} & 60.2 & \textbf{67.8} \\
bul\_Cyrl & \textbf{88.6} & \underline{88.3} & 87.3 & cat\_Latn & 86.6 & \textbf{86.9} & \underline{86.7} & ceb\_Latn & 50.1 & \underline{66.3} & \textbf{68.8} \\
ces\_Latn & \textbf{84.4} & \textbf{84.4} & 83.4 & cym\_Latn & \textbf{65.8} & 64.7 & \underline{65.3} & dan\_Latn & \underline{90.3} & 90.1 & \textbf{90.7} \\
deu\_Latn & \textbf{88.4} & \underline{87.9} & 87.3 & ell\_Grek & \textbf{88.0} & \underline{83.9} & 82.1 & eng\_Latn & \textbf{96.3} & \underline{96.1} & \underline{96.1} \\
est\_Latn & \textbf{85.9} & 82.5 & \underline{83.7} & eus\_Latn & \textbf{71.2} & 61.1 & \underline{63.6} & fao\_Latn & 77.6 & \underline{89.1} & \textbf{89.4} \\
\bottomrule
    \end{tabular}
}
    \caption{F1 scores of baselines and \modelname on \textbf{POS}.}\label{pos}
\end{table*}



\end{document}